\documentclass[twoside]{article}

\usepackage[utf8]{inputenc} 
\usepackage[T1]{fontenc}    
\usepackage{hyperref}       
\usepackage{url}            
\usepackage{booktabs}       
\usepackage{amsfonts}       
\usepackage{nicefrac}       
\usepackage{microtype}      
\usepackage{xcolor}         
\usepackage{graphicx}
\usepackage{multirow}
\usepackage{bm}
\usepackage{amsmath}
\usepackage{lmodern}      
\usepackage{subfigure}
\usepackage{subcaption}
\usepackage{algorithm}
\usepackage{algorithmic}
\usepackage{enumitem}

\usepackage[accepted]{aistats2025}
\usepackage[round]{natbib}  

\begin{document}

%
\runningtitle{Adversarial Vulnerabilities in LLM4TS}

%

\twocolumn[

\aistatstitle{Adversarial Vulnerabilities in Large Language Models \\for Time Series Forecasting}

\aistatsauthor{ Fuqiang Liu\(^*\)\(^1\) \And Sicong Jiang\(^*\)\(^1\)  \And  Luis Miranda-Moreno\(^1\) \And Seongjin Choi\(^2\) \And Lijun Sun\(^1\) }
\vspace{0.3cm}
\aistatsaddress{ \(^1\)McGill University  \And \(^2\)University of Minnesota - Twin Cities}   

]

\begin{abstract}
 Large Language Models (LLMs) have recently demonstrated significant potential in time series forecasting, offering impressive capabilities in handling complex temporal data. However, their robustness and reliability in real-world applications remain under-explored, particularly concerning their susceptibility to adversarial attacks. In this paper, we introduce a targeted adversarial attack framework for LLM-based time series forecasting. By employing both gradient-free and black-box optimization methods, we generate minimal yet highly effective perturbations that significantly degrade the forecasting accuracy across multiple datasets and LLM architectures. Our experiments, which include models like LLMTime with GPT-3.5, GPT-4, LLaMa, and Mistral, TimeGPT, and TimeLLM show that adversarial attacks lead to much more severe performance degradation than random noise, and demonstrate the broad effectiveness of our attacks across different LLMs. The results underscore the critical vulnerabilities of LLMs in time series forecasting, highlighting the need for robust defense mechanisms to ensure their reliable deployment in practical applications. The code repository can be found at \href{https://github.com/JohnsonJiang1996/AdvAttack_LLM4TS}{\textcolor{blue}{Johnson/AdvAttackLLM4TS}}.
\end{abstract}

\section{INTRODUCTION}

Time series forecasting plays a pivotal role in numerous real-world applications, ranging from finance and healthcare to energy management and climate modeling. 
Accurately predicting temporal patterns in the data is crucial for informed decision-making in these domains \citep{liu2022universal,jiang2024empowering}.
Recently, Large Language Models (LLMs), originally designed for Natural Language Processing (NLP) tasks, have demonstrated remarkable potential in handling time series forecasting challenges \citep{gruver2024large,liu2024taming,liu2024timecma,tan2024language,jin2023time}. These models, including BERT \citep{devlin2018bert}, GPT \citep{brown2020language, achiam2023gpt}, LLaMa \citep{touvron2023llama} and their successors, leverage their powerful attention mechanisms and vast pre-training on diverse datasets to capture intricate and non-linear temporal dependencies, making them highly effective for complex forecasting tasks.

LLMs exhibit strong generalization capabilities across various types of time series data. Compared to traditional models like ARIMA \citep{kalpakis2001distance} and Exponential Smoothing \citep{gardner1985exponential}, as well as advanced deep learning models such as DNNs \citep{salinas2020deepar, oreshkin2019n,challu2023nhits}, and Transformer-based architectures \citep{zhou2021informer,wu2021autoformer,liu2023itransformer,zhou2022fedformer}, LLMs excel in modeling long-term dependencies and capturing non-linear patterns within temporal sequences. This has resulted in impressive forecasting accuracy across applications ranging from energy consumption predictions to weather forecasting \citep{jin2023large}. 

However, despite their success, the robustness and reliability of LLMs in real-world forecasting applications remain concerns, particularly their vulnerability to adversarial attacks is under-explored. Adversarial attacks introduce subtle, often imperceptible perturbations to input data, leading to significant and misleading changes in model predictions. While the susceptibility of machine learning models to such attacks has been well-explored in image processing and NLP domains \citep{xu2020adversarial, morris2020textattack,wei2018transferable}, there is a noticeable gap in research on their impact on LLMs used for time series forecasting. 

\begin{figure*}
    \centering    \includegraphics[width=0.9\linewidth]{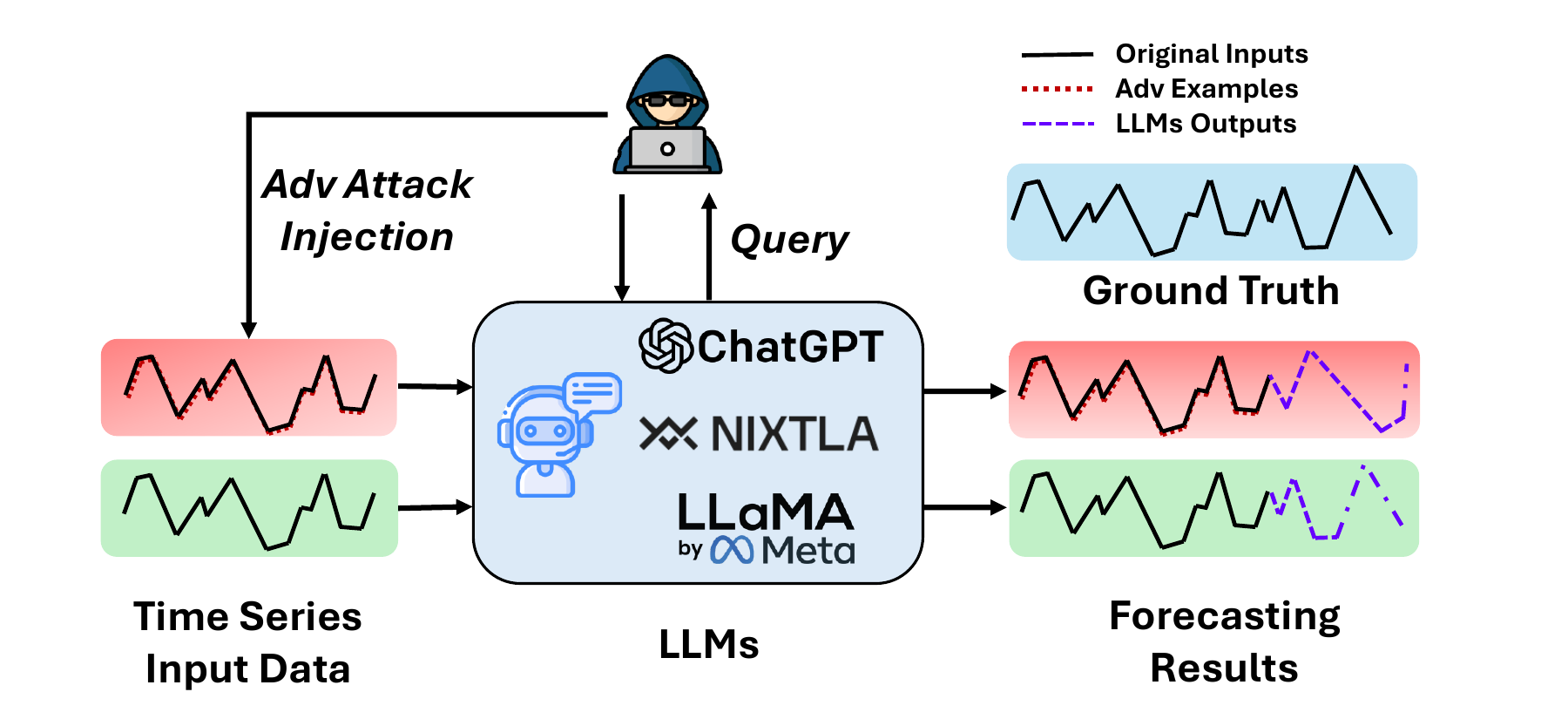}
    \caption{Adversarial Black-box Attack for LLMs in Time Series Forecasting.}
    \label{fig:schemetic}
\end{figure*}

While adversarial attacks and defenses for deep neural networks have been extensively studied across various domains \citep{madry2017towards}, executing adversarial attacks against LLMs in time series forecasting presents two significant challenges. First, to prevent information leakage, we cannot use ground truth values (i.e., future time steps) when attacking forecasting models. Second, LLMs must be treated as strict black-box systems due to the difficulty of accessing their internal workings and parameters.

In this paper, we address this gap by proposing a gradient-free black-box attack that transforms the output of LLM-based forecasting models into a random walk, while investigating the vulnerabilities of large language models in time series forecasting. As depicted in Figure~\ref{fig:schemetic}, we demonstrate that even minimal attack perturbations can cause substantial deviations in LLMs' predictions. We evaluate three forms of LLM applications for time series forecasting, encompassing six sub-models as well as two non-LLM models, across five datasets from various real-world domains. Our findings reveal that LLMs, despite their advanced architectures, are indeed susceptible to adversarial manipulations in time series domain, leading to unstable and inaccurate forecasts. This underscores the urgent need to develop more robust LLMs that can withstand such attacks, ensuring their reliability in real-world applications. 


In conclusion, this study contributes to the ongoing discourse on the robustness of LLMs by revealing their vulnerabilities to adversarial attacks in time series forecasting. Our findings underscore the critical need to address these vulnerabilities to ensure that LLM-based forecasting models are not only accurate but also resilient, thereby enhancing their practical utility in high-stakes applications. 

\section{RELATED WORK}
\subsection{Adversarial Attacks on Forecasting}
Adversarial attacks in time series forecasting have emerged as a crucial area of research, exposing vulnerabilities in forecasting models. Unlike adversarial studies in static domains, such as object recognition or time series classification, adversarial attacks on time series forecasting cannot leverage ground truth data for perturbation generation due to the risk of information leakage \citep{liu2022practical}. To address this challenge, surrogate techniques have been adopted \citep{liu2021spatially}, which bypass the need for labels, as is done in traditional adversarial attack methods like the Fast Gradient Sign Method \citep{goodfellow2014explaining}. Several studies have treated forecasting models as white-box systems to investigate the effects of adversarial attacks on commonly used models in time series forecasting, such as ARIMA, LSTMs, and Transformer-based models \citep{liu2022robust,liu2023adversarial}. These studies demonstrate that even small perturbations can severely impact these models, resulting in inaccurate forecasts. However, evaluating the vulnerability of LLM-based forecasting presents a significant challenge, as internal access is typically restricted, requiring these models to be treated as black-box systems.

\subsection{Adversarial Attacks on LLMs}

Adversarial attacks on LLMs have gained increasing attention, focusing on how slight manipulations can significantly alter their outputs. These attacks are often classified into prompt-based attacks, token-level manipulations, gradient-based attacks, and embedding perturbations.
\begin{itemize}[noitemsep,topsep=0pt] \item Jailbreak Prompting \citep{yu2024don,wei2024jailbroken}: Crafted prompts that bypass LLM guardrails, inducing unintended or harmful outputs by exploiting unconventional phrasing.

\item Prompt Injection \citep{greshake2023not,xue2024trojllm,liu2024automatic}: Adversarial instructions embedded into benign prompts to manipulate LLM responses, highlighting their vulnerability to prompt manipulation.

\item Gradient Attacks \citep{madry2017towards,guo2021gradient}: Using internal model parameters, attackers apply gradient-based methods to perturb inputs, significantly altering outputs with minimal changes.


\item Embedding Perturbations \citep{schwinn2024soft,singh2024robustness}: Subtle changes to input embeddings disrupt LLM's internal representations, leading to erroneous outputs with minimal visible input alterations. \end{itemize}
While extensive research has been conducted on attacks against LLMs at various levels, most of these focus on text-based manipulations. However, there's a significant gap in understanding how LLMs perform in non-textual tasks, particularly time series forecasting. In language tasks, attacks typically manipulate static text inputs, such as words or prompts, to exploit the LLM's understanding and induce specific outputs. However, time series forecasting involves dynamic, evolving data points, requiring attackers to introduce perturbations that maintain the sequence's natural flow and coherence.

\section{MANIPULATING LLM-BASED TIME SERIES FORECASTING}\label{sec:adversarial attack}

\subsection{Formulations of LLM-based Forecasting}
LLMs have shown promising performance in time series forecasting by leveraging their ability to perform next-token prediction, a technique originally developed for text-based tasks \citep{gruver2024large,jin2023time}. A typical LLM-based time series forecasting model, denoted as $f(\cdot)$, consists of two primary components: an embedding or tokenization module that encodes the time series data into a sequence of tokens, and a pre-trained LLM that autoregressively predicts the subsequent tokens. The embedding module translates the raw time series into a format suitable for the LLM, while the LLM captures the temporal dependencies and generates predictions based on its learned representations.

Let $\mathbf{X}_t \in \mathbb{R}^d$  denote $d$-dimensional time series at time $t$, where $x_{i,t}=[\mathbf{X}_t]_i$ represents the observation of the $i$-th component of the time series. Given a sequence of recent $T$ historical observations $\mathbf{X}_{t-T+1:t}$, a forecasting model, $f(\cdot)$, is employed to predict the future values for the subsequent $\tau$ time steps. The prediction is formulated as:
\begin{equation}
\hat{\mathbf{Y}}_{t+1:t+\tau} = f\left(\mathbf{X}_{t-T+1:t}\right),    
\end{equation}
where $\hat{\mathbf{Y}}_{t+1:t+\tau}$ denotes the predicted future values and $\mathbf{Y}_{t+1:t+\tau}$ represents the corresponding ground truth values. It is important to note that the prediction horizon is typically less than or equal to the historical horizon, i.e., $\tau \le T$.

\subsection{Threat Model}
Our objective is to deceive an LLM-based time series forecasting model into producing anomalous outputs that deviate significantly from both its normal predictions and the corresponding ground truth, through the introduction of imperceptible perturbations. This adversarial attack problem can be framed as an optimization task as follows:
\begin{equation}~\label{equ:foreattack_1}
\begin{split}
\max_{\rho_{t-T+1:t}}~& \mathcal{L}\left(f\left(\mathbf{X}_{t-T+1:t}+\bm{\rho}_{t-T+1:t} \right), \mathbf{Y}_{t+1:t+\tau}\right)\\
\text{s.t.}~&\ \left\|\rho_i \right\|_p \le\epsilon, i\in\left[t-T+1,t\right],   
\end{split}
\end{equation}
where $\mathbf{X}_{t-T+1:t}$ denotes the clean input, $\mathbf{Y}_{t+1:t+\tau}$ denotes the true future values, and $\bm{\rho}_{t-T+1:t}$ denotes the adversarial perturbations.

The loss function $\mathcal{L}$ quantifies the discrepancy between the model's output and the ground truth, while $\epsilon$ constrains the magnitude of the perturbations under the $\ell_p$-norm, ensuring that the adversarial attack remains imperceptible.

Since the true future values $\mathbf{Y}_{t+1:t+\tau}$ are typically inaccessible in practical time series forecasting, they are replaced with the predicted values $\hat{\mathbf{Y}}_{t+1:t+\tau}$ generated by the forecasting model. Consequently, Eq.~\ref{equ:foreattack_1} is reformulated as
\begin{equation}~\label{equ:foreattack_2}
\begin{split}
\max_{\bm{\rho}_{t-T+1:t}}~& \mathcal{L}\left(f\left(\mathbf{X}_{t-T+1:t}+\bm{\rho}_{t-T+1:t} \right), \hat{\mathbf{Y}}_{t+1:t+\tau}\right)\\
\text{s.t.}~&\ \left\|\rho_i \right\|_p \le\epsilon, i\in\left[t-T+1,t\right].   
\end{split}
\end{equation}

In practical applications, accessing the full set of detailed parameters of an LLM is typically infeasible, which forces the attacker to treat the target model as a black-box system. Additionally, acquiring the complete training dataset is impractical, meaning the attacker lacks access to this information as well. The attacker’s capabilities can be summarized as follows:
\begin{itemize}[noitemsep,topsep=0pt]
    \item \textbf{no access to the training data}, 
    \item \textbf{no access to internal information of the LLM-based forecasting model}, 
    \item \textbf{no access to ground truth}, 
    \item \textbf{the ability to query the target model}.
\end{itemize}

The threat model for adversarial attacks against LLM-based time series forecasting underscores the complexity of this task. Unlike attacks on LLMs in static applications, the attacker here cannot leverage labels for crafting attacks. Furthermore, compared to attacks on non-LLM forecasting models, the internal details of LLMs are strictly inaccessible, prohibiting the use of white-box attack techniques. This restriction highlights the increased challenge of developing effective adversarial attacks in this context.



\section{DIRECTIONAL GRADIENT APPROXIMATION}
Since the attacker has no access to the internal parameters of the LLM, it is not feasible to compute gradients and use them to solve the optimization problem presented in Eq.~\ref{equ:foreattack_2}. This results in a gradient-free optimization problem. To address this, we propose a gradient-free optimization approach, referred to as targeted attack with \textbf{D}irectional \textbf{G}radient \textbf{A}pproximation (DGA), aimed at generating perturbations that can effectively deceive LLM-based time series forecasting models.

We first adjust our objective to focus on misleading the forecasting model into producing outputs that closely resemble an anomalous sequence, rather than simply deviating from its normal predictions. Accordingly, the optimization problem in Eq.~\ref{equ:foreattack_2} is reformulated as
\begin{equation}~\label{equ:foreattack_3}
\begin{split}
\min_{\bm{\rho}_{t-T+1:t}}~& \mathcal{L}\left(f\left(\mathbf{X}_{t-T+1:t}+\bm{\rho}_{t-T+1:t} \right), \mathcal{Y}\right)\\
\text{s.t.}~&\ \left\|\rho_i \right\|_p \le\epsilon, i\in\left[t-T+1,t\right],   
\end{split}
\end{equation}
where $\mathcal{Y}$ represents the targeted anomalous time series.

Supposing $\bm{\theta}_{t-T+1:t}$ denote a random small signal, the gradient, $\bm{g}_{t-T+1:t}$, which approximates the direction from the normal output to the targeted anomalous output, can be expressed as
\begin{equation}\label{equ:gradient}
\begin{split}
&\bm{g}_{t-T+1:t}=\\ &\frac{\mathcal{L}\left(\mathcal{Y}-f\left(\mathbf{X}_{t-T+1:t}+\bm{\theta}_{t-T+1:t} \right)\right)-\mathcal{L}\left(\mathcal{Y}-f\left(\mathbf{X}_{t-T+1:t}\right)\right)}{\bm{\theta}_{t-T+1:t}}.     
\end{split}
\end{equation}
Supposing $\ell_1$-norm is applied in Eq.~\ref{equ:foreattack_3}, the magnitude of the perturbation is strictly constrained to be imperceptible. The perturbation, $\bm{\rho}_{t-T+1:t}$, can be computed from the approximated gradient, and the temporary adversarial example, $\mathbf{X'}_{t-T+1:t}$, is generated as
\begin{equation}\label{equ:perturbation}
\begin{split}
\mathbf{X'}_{t-T+1:t} &= \mathbf{X}_{t-T+1:t}+\bm{\rho}_{t-T+1:t}\\&=\mathbf{X}_{t-T+1:t}+\epsilon \cdot \text{sign}\left(\bm{g}_{t-T+1:t}\right),    
\end{split}
\end{equation}
where $\text{sign}\left(\cdot\right)$ denotes the signum function.

A time series forecasting model that produces Gaussian White Noise (GWN) as its output is considered to generate an anomalous prediction. Consequently, GWN can be utilized as the target sequence in Eq.~\ref{equ:perturbation}, formulated as $\mathcal{Y}\sim \mathcal{N}\left(\mu, \sigma\right)$, where $\mu$ and $\sigma$ represent the mean and the standard deviation, respectively. Empirically, the mean and standard deviation of the input data can be used to generate GWN. This results in a situation where a temporally correlated time series is misleadingly predicted as independent and identically distributed (i.i.d.) noise. This approach highlights the model's inability to preserve temporal correlations when subjected to adversarial perturbations, thereby reinforcing the effectiveness of the adversarial attack.

\section{EXPERIMENTS}
\subsection{Datasets}\label{sec:datasets}
To evaluate the proposed DGA and gain a further understanding of the vulnerability of LLM-based forecasting, We conducted experiments using five widely recognized real-world datasets that cover a broad range of time series forecasting tasks:

\begin{itemize}[noitemsep,topsep=0pt]
    \item \textbf{ETTh1 and ETTh2 (Electricity Transformer Temperature Hourly) \citep{zhou2021informer}}: These datasets consist of two years of hourly recorded data from electricity transformers, capturing temperature and power consumption variables. 
    
    \item \textbf{IstanbulTraffic \citep{gruver2024large}}: This dataset contains hourly measurements of road traffic volumes across different sensors. It captures temporal dependencies related to traffic patterns, which are dynamic and fluctuating time series.

    \item \textbf{Weather \citep{wu2022timesnet}}: This dataset comprises meteorological data, including variables such as temperature, humidity, and wind speed, recorded hourly. It provides a challenging forecasting task due to the inherent variability and complexity of weather patterns.

    \item  \textbf{Exchange \citep{lai2018modeling}}: This dataset consists of daily exchange rates from eight foreign countries—Australia, the United Kingdom, Canada, Switzerland, China, Japan, New Zealand, and Singapore---covering the period from 1990 to 2016.
\end{itemize}
    
These diverse datasets allow us to evaluate the adversarial robustness of LLMs across different types of temporal dynamics and forecasting challenges. In our experiments, 50\% of the data is used for training, while the remaining data is split evenly: 25\% for validation and 25\% for testing. We use a 96-step historical time window as input to the forecasting model, which predicts the subsequent 48-step future values. It should be noted that the attacker does not access either the training or validation part.

\subsection{Target Models}
To assess the impact of adversarial attacks on LLMs for time series forecasting, we selected three state-of-the-art LLM-based forecasting models as baselines, which together represent three common forms of LLM application for time series tasks:
\begin{itemize}[noitemsep,topsep=0pt]
\item \textbf{TimeGPT} \citep{garza2023timegpt}: A large model specifically pre-trained with a vast amount of time series data. It uses advanced attention mechanisms and temporal encoding to capture complex patterns in sequential data, making it a leading LLM designed explicitly for time series forecasting. Its pre-training, which is conducted from scratch using vast amounts of time series data, allows it to serve as a robust and versatile tool for a wide range of time-dependent applications. 
\item \textbf{LLMTime} \citep{gruver2024large}: This model treats time series forecasting as a next-token prediction task, using LLM architectures like GPT and LLaMa. By converting time series data into numerical sequences, LLMTime enables these models to apply their sequence prediction strengths to time series. To test the robustness of our adversarial attacks, we experimented with base models including GPT-3.5, GPT-4, LLaMa, and Mistral, assessing their resilience when adapted from natural language processing to time series forecasting.
\item \textbf{TimeLLM} \citep{jin2023time}: This model presents a novel approach for time series forecasting by adapting LLMs by reprogramming input time series into textual representations that are more compatible with LLMs, allowing the models to perform time series forecasting tasks without altering their pre-trained structures. The key innovation is the Prompt-as-Prefix (PaP) technique, which augments input context to guide the LLM in transforming reprogrammed data into accurate forecasts.
\item \textbf{TimeNet} \citep{wu2022timesnet} and \textbf{iTransformer} \citep{liu2023itransformer}are two supervised forecasting models that leverage the attention-based architecture of transformers, enabling them to capture long-term dependencies in time series data effectively. These models provide strong performance in time series forecasting tasks, and in this study, they are used in the performance and robustness comparison between LLM-based forecasting models with non-LLM models. 
\end{itemize}

Both TimeGPT and LLMTime operate as zero-shot forecasters, making predictions for each time series without prior exposure to the dataset. In contrast, TimeLLM is fine-tuned using 10\% of each dataset for every task. On the other hand, the two non-LLM forecasters, TimeNet and iTransformer, are trained using the full available training data for each dataset.

While we selected only three baseline LLM-based models for this study, the setup encompasses the primary approaches to LLM-based time series forecasting: pre-training a large model specifically for time series data (e.g., TimeGPT), leveraging well-developed general-purpose language models (e.g., LLMTime), and fine-tuning language models from other domains for time series forecasting (e.g., TimeLLM). This comprehensive selection provides a representative overview of the key strategies in adapting LLMs for time series tasks. 


\subsection{Experimental Procedures}
We designed a series of experiments to evaluate the vulnerability of the baseline LLM models to adversarial attacks. For each model and dataset combination, we conducted the following procedures: (i) we applied targeted perturbations to the input data, carefully maintaining the overall structure of the original time series while subtly altering the data to mislead the LLMs' forecasting predictions; (ii) we introduced GWN with the same perturbation intensity; (iii) forecasting accuracy was measured using Mean Absolute Error (MAE) and Mean Squared Error (MSE), which allowed us to quantify the performance degradation caused by adversarial attacks compared to Gaussian noise.

\begin{table*}[!t]
\centering
\caption{Results for univariate time series forecasting with a consistent input length of 96 and an output length of 48 across all models and datasets. A lower MSE or MAE indicates better prediction performance. The perturbation scale is set to 2\% of the mean value of each dataset. Bold text highlights the worst performance for each dataset and model combination.}
\resizebox{\textwidth}{!}{
\begin{tabular}{l|cc|cc|cc|cc|cc|cc|cc|cc}
\toprule
\multirow{2}{*}{Models} & \multicolumn{2}{c|}{LLMTime} & \multicolumn{2}{c|}{LLMTime} & \multicolumn{2}{c|}{LLMTime} & \multicolumn{2}{c|}{LLMTime} & \multicolumn{2}{c|}{Time-LLM} & \multicolumn{2}{c|}{TimeGPT} &\multicolumn{2}{c|}{iTransformer} & \multicolumn{2}{c}{TimesNet}\\
                        & \multicolumn{2}{c|}{w/ GPT-3.5} & \multicolumn{2}{c|}{w/ GPT-4} & \multicolumn{2}{c|}{w/ LLaMa 2} & \multicolumn{2}{c|}{w/ Mistral} & \multicolumn{2}{c|}{w/ GPT-2} & \multicolumn{2}{c|}{(2024)} &\multicolumn{2}{c|}{(2024)} &\multicolumn{2}{c}{(2023)} \\\midrule
                        Metrcis& MSE & MAE  & MSE & MAE  & MSE & MAE  & MSE & MAE  & MSE & MAE & MSE & MAE & MSE & MAE & MSE & MAE \\
\midrule
ETTh1                   & 0.073    & 0.213 &   0.071  & 0.202 &  0.086   & 0.244 &    0.097 & 0.274 &   0.089 & 0.202  & 0.059   & 0.192  &0.071 &0.218 &0.073&0.202\\
w/ GWN                  &   0.077  & 0.219 & 0.076    & 0.213 & 0.087    & 0.237 &   0.094  & 0.291 &    \textbf{0.102} & 0.231 & 0.059 & 0.193&  0.072& 0.216 &0.074&0.202\\
w/ DGA                  & \textbf{0.085}    & \textbf{0.249} &    \textbf{0.083} & \textbf{0.232} &    \textbf{0.091} & \textbf{0.251} &   \textbf{0.098}  & \textbf{0.295} &    0.099 & \textbf{0.248} & \textbf{0.060} & \textbf{0.198} &  \textbf{0.075} & \textbf{0.226} &\textbf{0.081}&\textbf{0.213}\\
\midrule
ETTh2                   &   0.263  & 0.372 &  0.155 & 0.267 &    0.237 & 0.373 &    0.277 & 0.492 &   0.238 & 0.361 & 0.161  & 0.297 & 0.171& 0.296 & 0.166&0.316\\
w/ GWN                  &    0.263 & 0.342 &    0.175 & 0.303 &   0.231  & \textbf{0.429} &   0.346  & 0.505 &  0.235   & 0.355 &  0.160   & 0.301 &\textbf{0.181}&0.302 &0.166&0.316\\
w/ DGA                  &    \textbf{0.275} & \textbf{0.408} &   \textbf{0.201}  & \textbf{0.327} &    \textbf{0.257} & 0.425 &     \textbf{0.356} & \textbf{0.554} & \textbf{0.302} & \textbf{0.441} &  \textbf{0.171}   & \textbf{0.312}  &0.179&\textbf{0.308} & \textbf{0.169}&\textbf{0.321}\\
\midrule
IstanbulTraffic          &  0.837 & 0.844 &   0.805  & 0.779 &   0.891  & 1.005 &   0.826  & 0.973 &   0.995& 1.013 & 1.890  & 1.201& 1.081& 0.995& 1.095 &1.022\\
w/ GWN                  &  0.882   & 0.908 &    0.883 & 0.864 &   0.917  & 1.063 &  1.054   & 1.031 & 1.123& 1.221 & 1.848  & 1.204 & \textbf{1.103} & 1.015 &1.103&1.035\\
w/ DGA                  &    \textbf{0.955} & \textbf{1.073} &   \textbf{1.417}  & \textbf{1.214} &   \textbf{0.994}  & \textbf{1.083} &   \textbf{1.744}  & \textbf{1.217} & \textbf{1.161} & \textbf{1.328} &    \textbf{1.918} & \textbf{1.218} & 1.097&\textbf{1.034} &\textbf{1.155}&\textbf{1.047} \\
\midrule
Weather                 &    0.005 & 0.051 &    0.004 & 0.048 &  0.008   & 0.072 &  0.006   & 0.057 &   0.004  & 0.034 &  0.004  & 0.043  & 0.005&0.053&0.003&0.042\\
w/ GWN                  &  0.005   & 0.053 &   0.005  & 0.051 &  0.008   & 0.074 &   \textbf{0.007}  & \textbf{0.066} &  0.004  & 0.033 &  0.004  & 0.043 & \textbf{0.006}&0.063&0.003&0.042\\
w/ DGA                  &   \textbf{0.006}  & \textbf{0.063} &   \textbf{0.006}  & \textbf{0.061} &   \textbf{0.009}  & \textbf{0.079} &    \textbf{0.007} & 0.062 & \textbf{0.005}  & \textbf{0.052} & \textbf{0.006}  & \textbf{0.071}&\textbf{0.006}&\textbf{0.065}&\textbf{0.004}&\textbf{0.045}\\
\midrule
Exchange                &    0.038 & 0.146 &   0.040  & 0.152 &   0.043  & 0.167&   0.151  & 0.274 &   0.056  & 0.188 &  0.256  & 0.368 &0.034 &0.151 & 0.056&0.184\\
w/ GWN                  &    0.042 & 0.179 &    0.046 & 0.182 &   0.050  & 0.185 & 0.160    & 0.298 & 0.059  & 0.194 &  0.329  & 0.413 & 0.044&0.166 & \textbf{0.065}&\textbf{0.195}\\
w/ DGA                  &  \textbf{0.058}   & \textbf{0.224} &   \textbf{0.068}  & \textbf{0.199} &   \textbf{0.069}  & \textbf{0.213} &   \textbf{0.219}  & \textbf{0.303} &    \textbf{0.077} & \textbf{0.256} & \textbf{0.578} & \textbf{0.556}&\textbf{0.049}&\textbf{0.178} &0.062&0.194\\
\bottomrule
\end{tabular}
}
\label{table:overall}

\end{table*}

\begin{figure*}[!h]
\centering
\subfigure[LLMTime w/ GPT-4]{
    \centering
    \includegraphics[width = 0.22\textwidth]{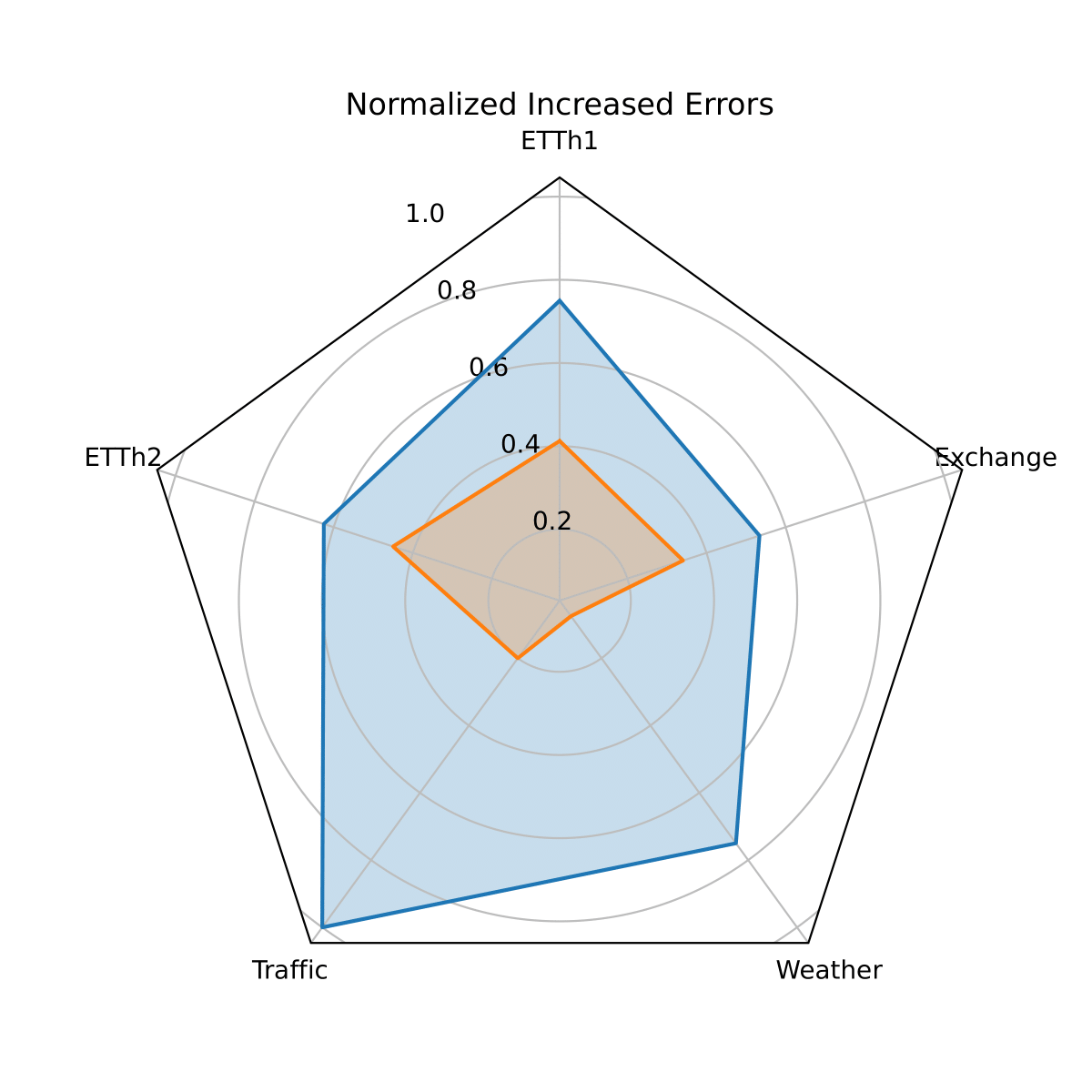}
}
\subfigure[TimeGPT]{
    \centering
    \includegraphics[width = 0.22\textwidth]{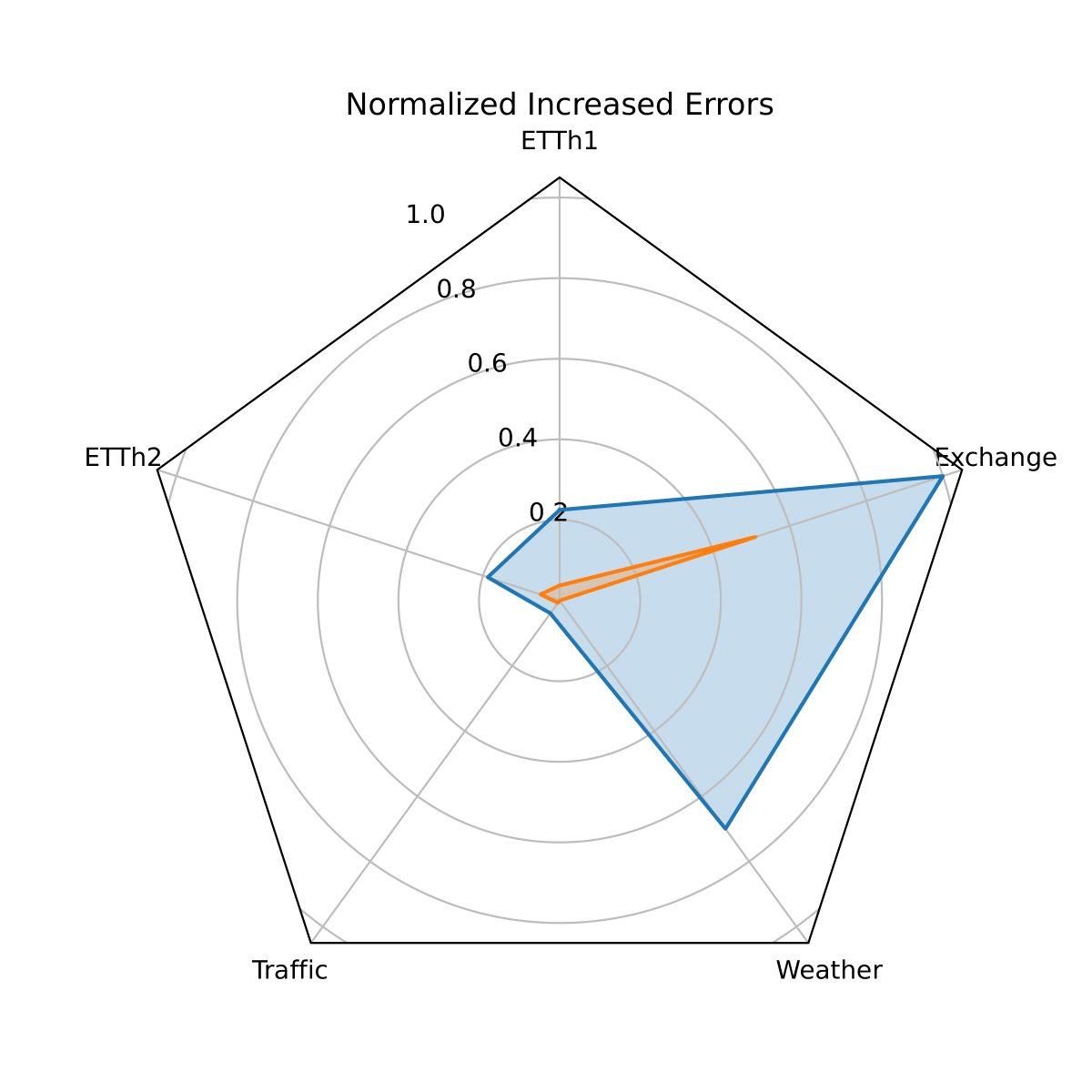}
}
\subfigure[iTransformer]{
    \centering
    \includegraphics[width = 0.22\textwidth]{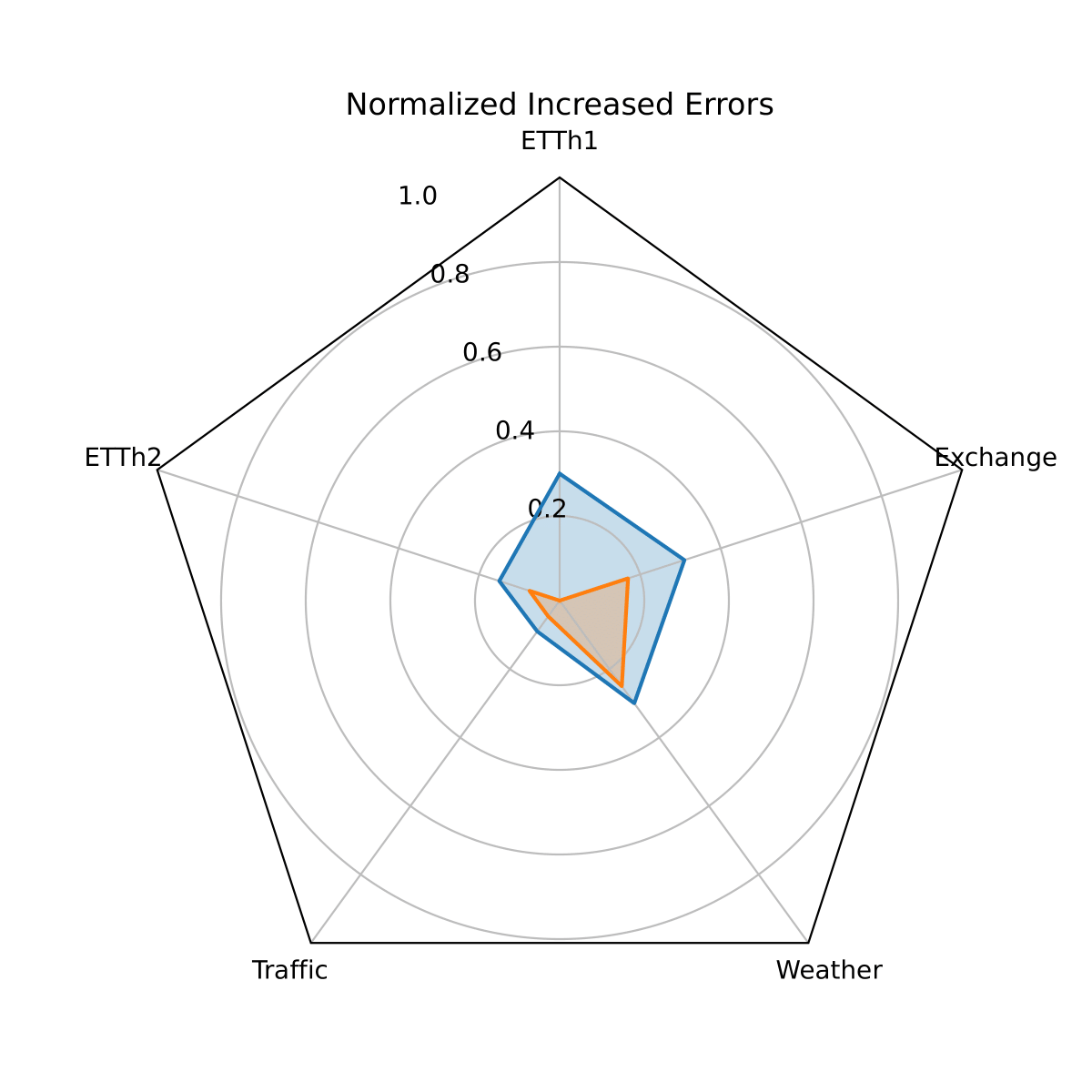}
}
\subfigure[TimeNet]{
    \centering
    \includegraphics[width = 0.22\textwidth]{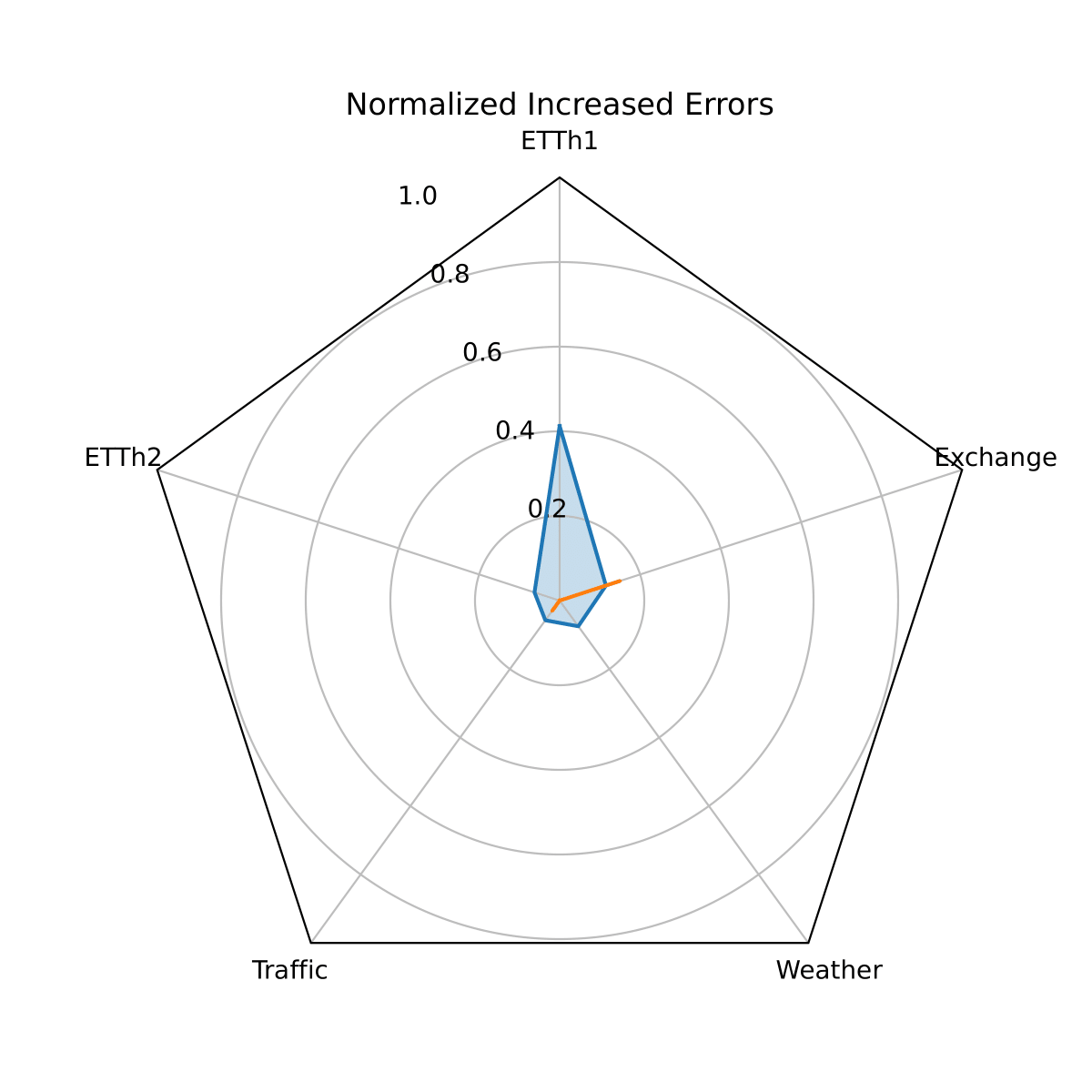}
}
\caption{Robustness comparison between LLM-based forecasting models and lighter models. These figures highlight each model's relative robustness across various datasets. The blue and orange shaded areas represent the normalized increase in MAE for each model under the influence of DGA and GWN perturbations, respectively. A larger shaded area indicates greater vulnerability to perturbations. }
\label{fig:robustness comparison}
\end{figure*}

\begin{figure*}[!t]
\centering
\subfigure[ETTh1, forecast: 02/13-02/15]{
    \centering
    \includegraphics[width = 0.45\textwidth]{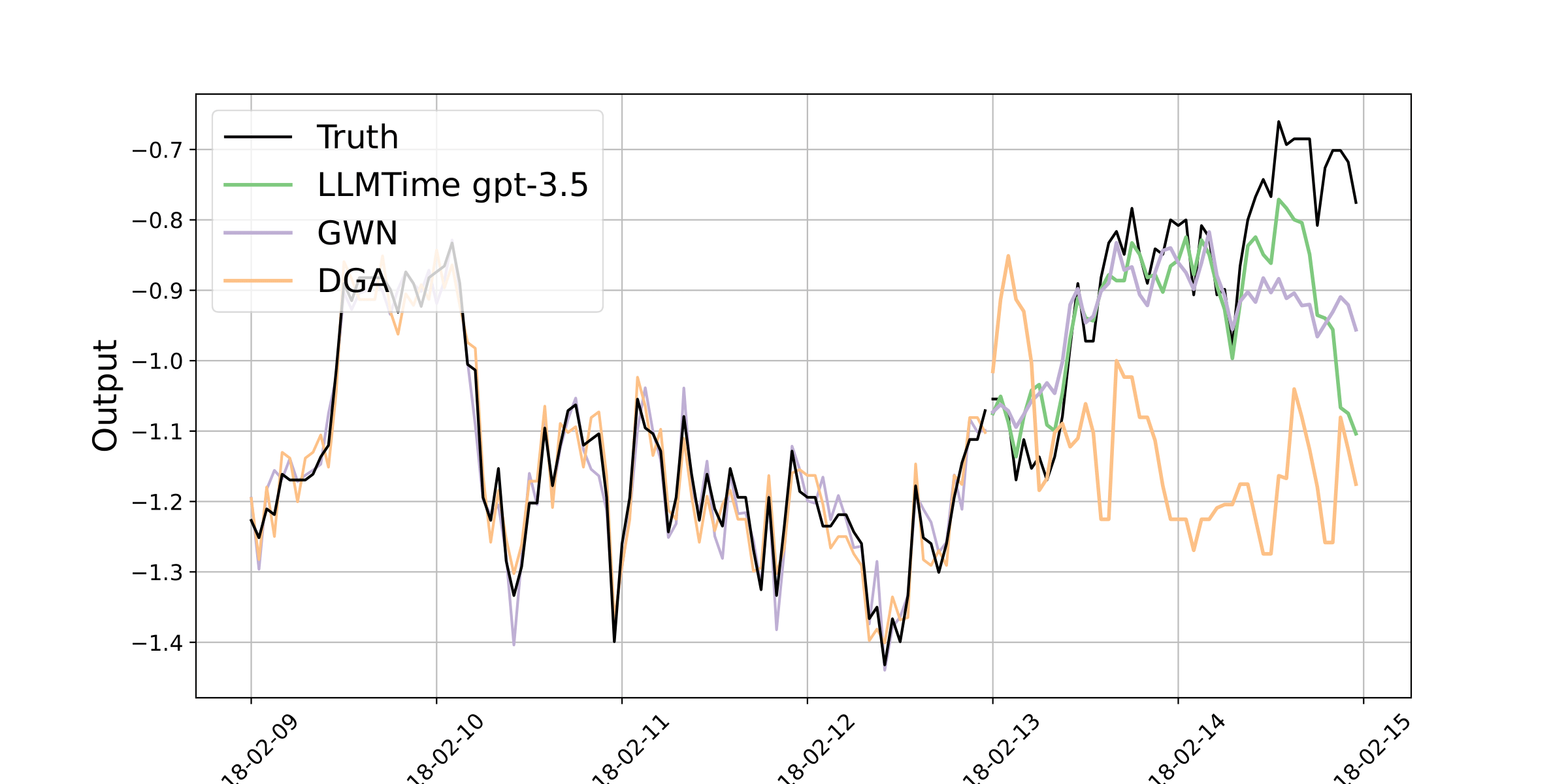}
}
\subfigure[ETTh1, input bias and prediction error]{
    \centering
    \includegraphics[width = 0.45\textwidth]{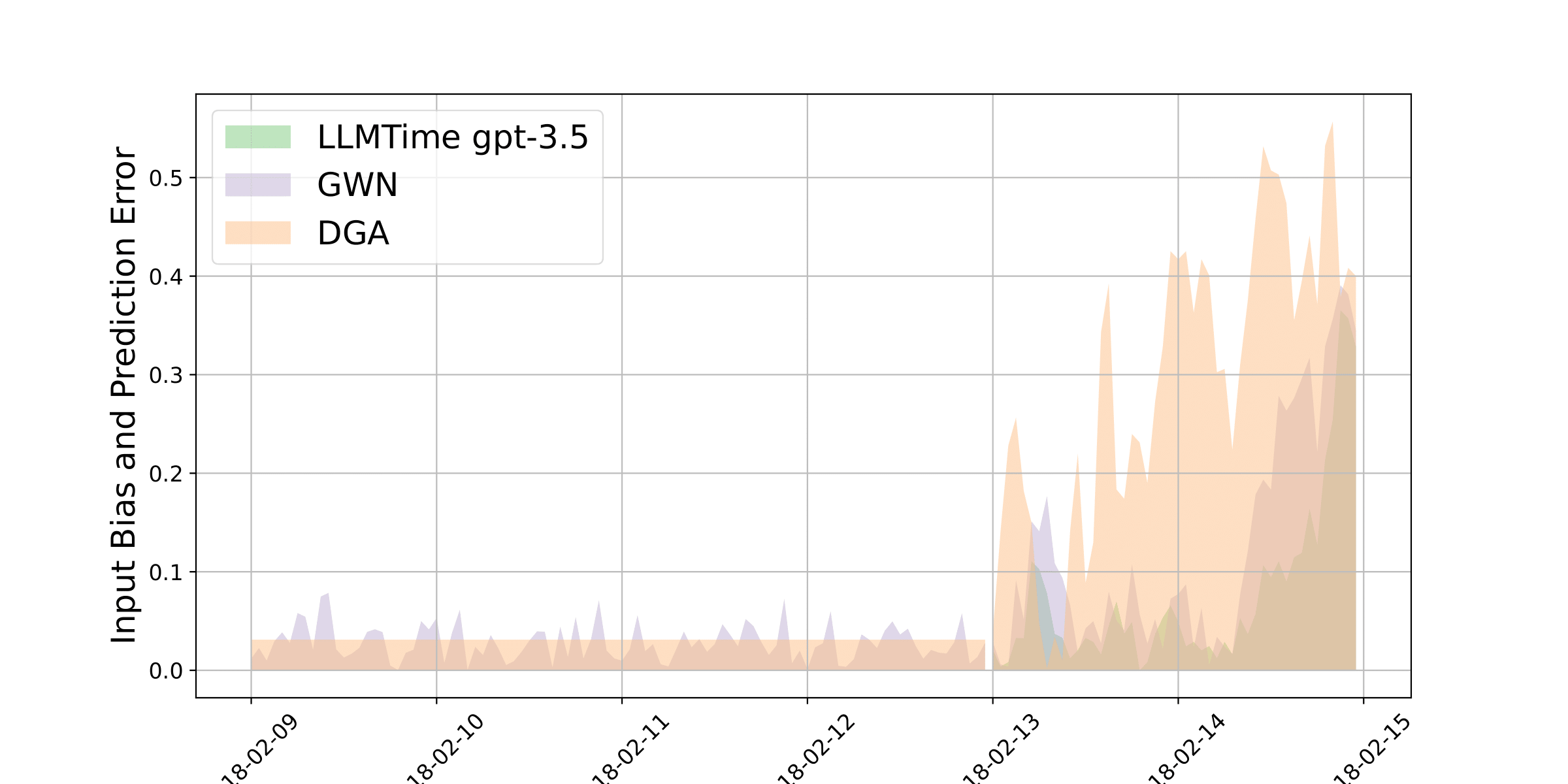}
}\\
\subfigure[ETTh2, forecast: 10/28-10/30]{
    \centering
    \includegraphics[width = 0.45\textwidth]{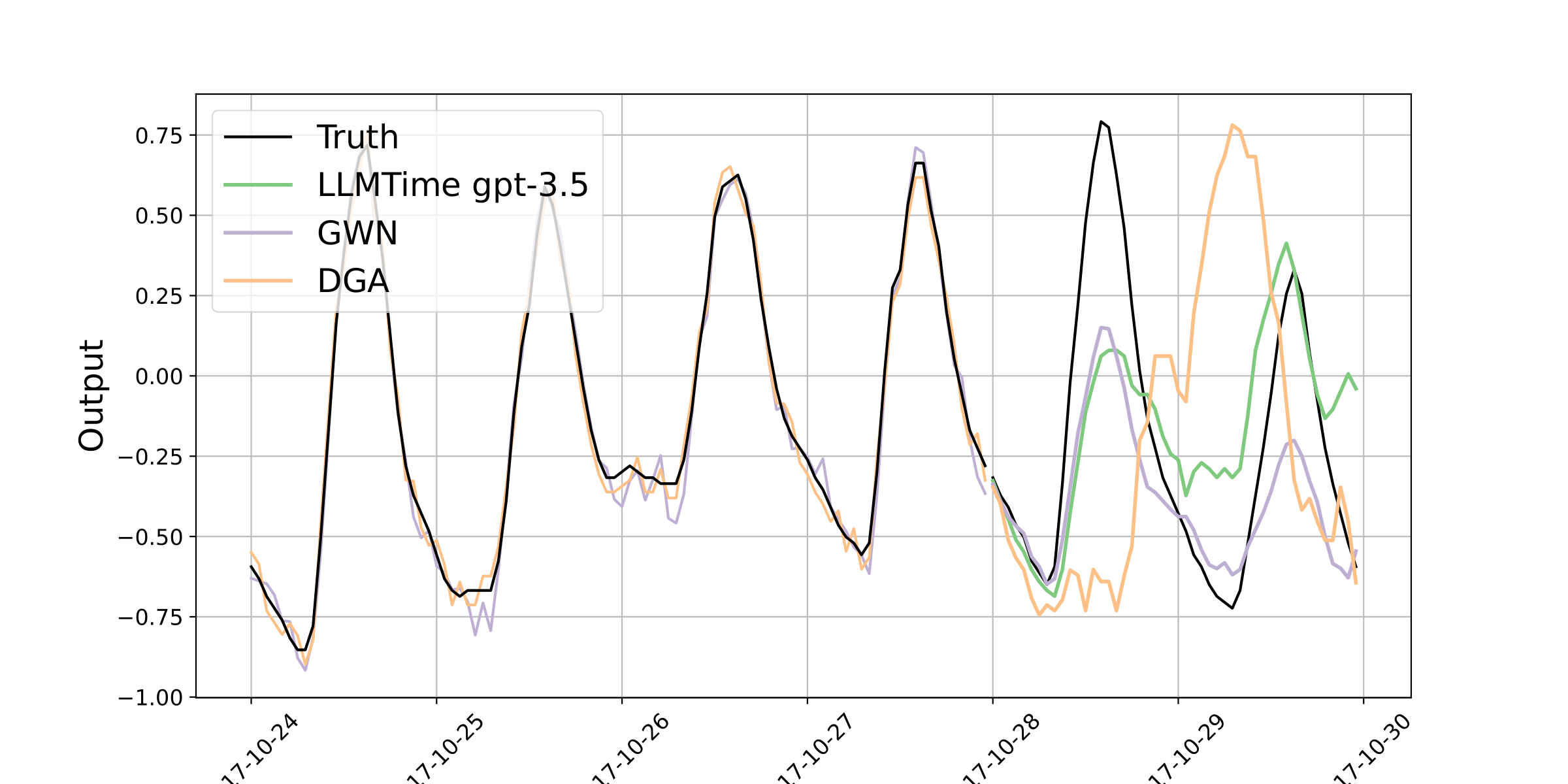}
}
\subfigure[ETTh2, input bias and prediction error]{
    \centering
    \includegraphics[width = 0.45\textwidth]{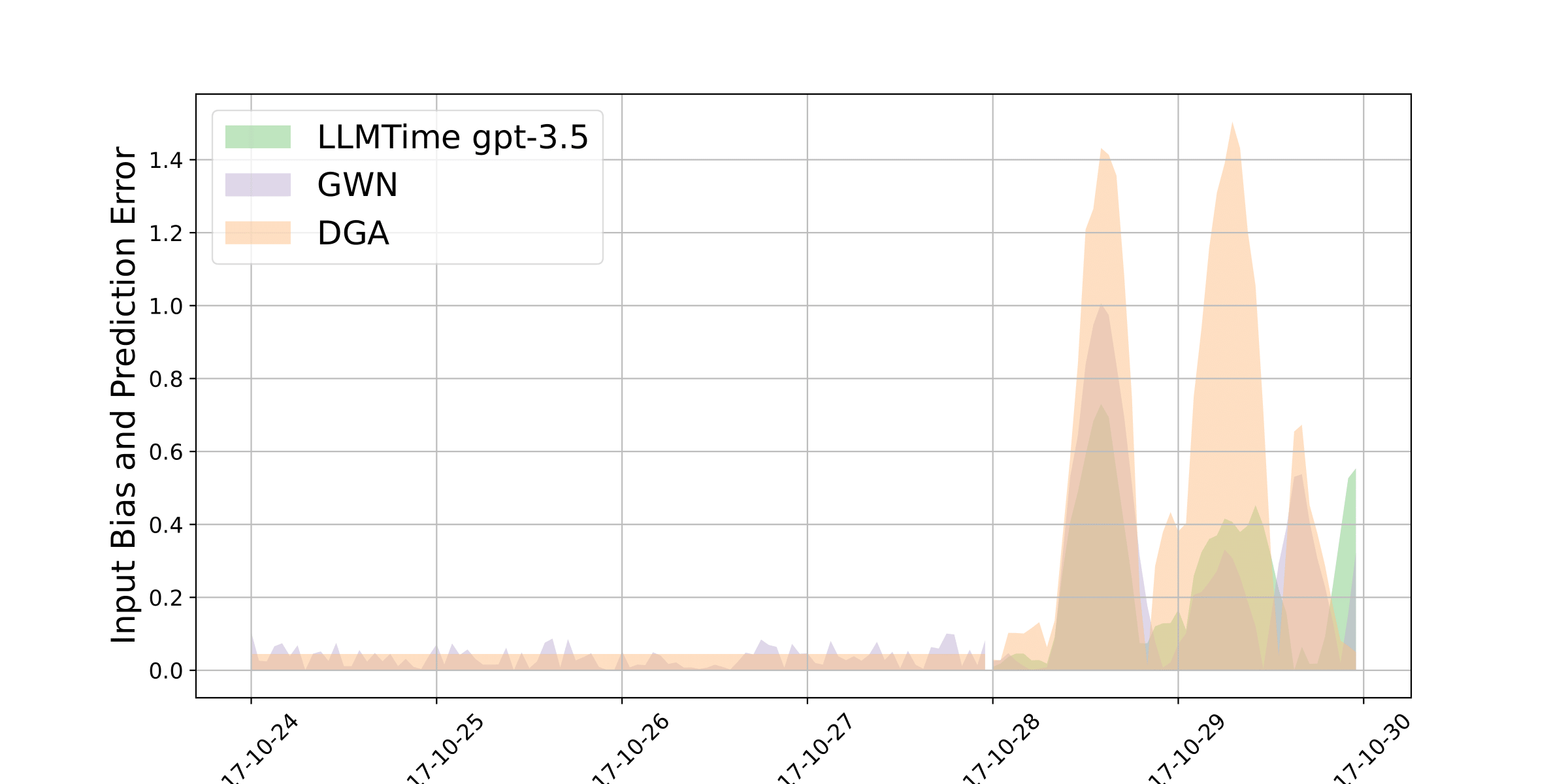}
}\\
\subfigure[weather, forecast: 3/21/16pm-3/22/12am]{
    \centering
    \includegraphics[width = 0.45\textwidth]{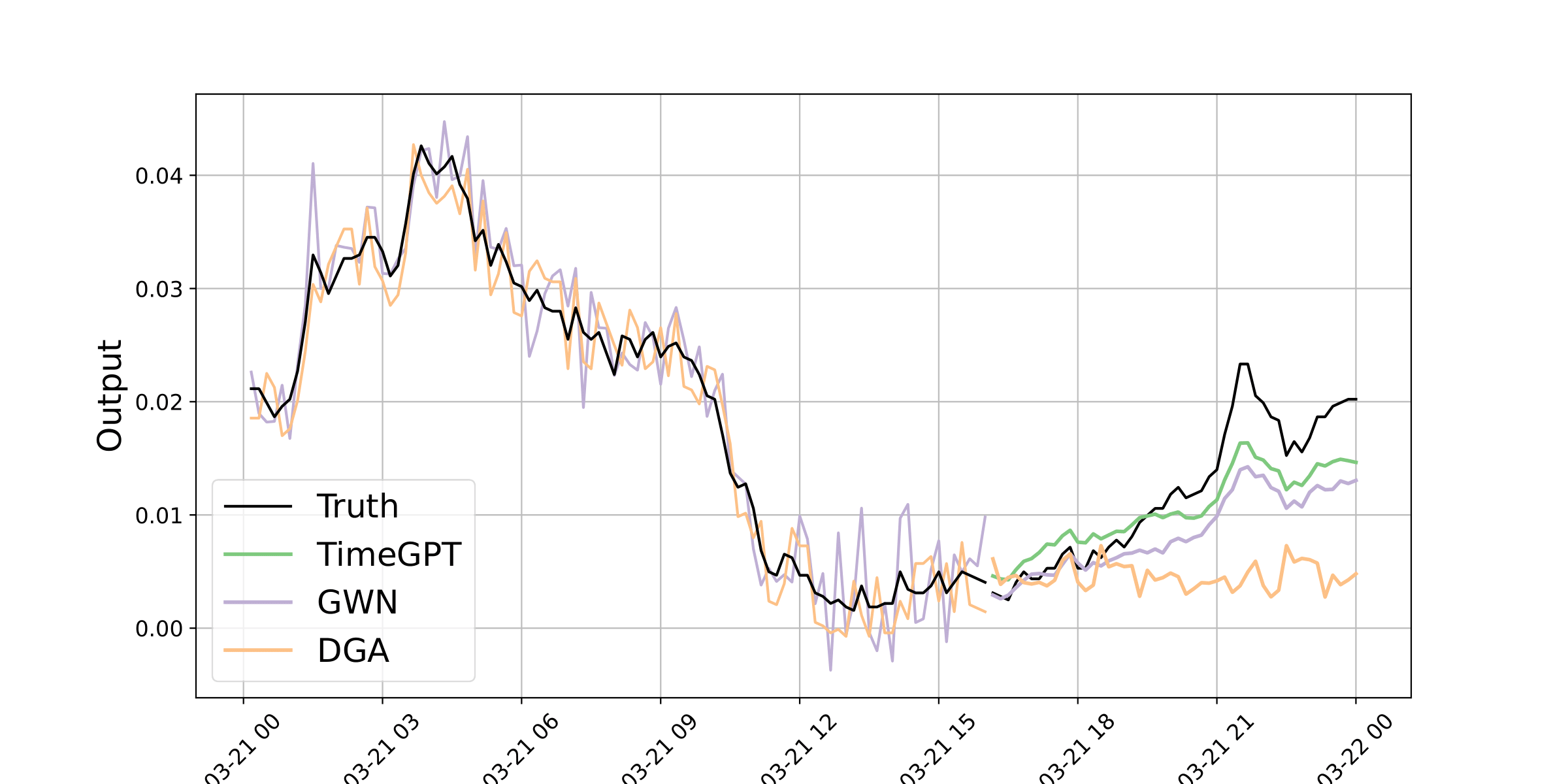}
}
\subfigure[weather, input bias and prediction error]{
    \centering
    \includegraphics[width = 0.45\textwidth]{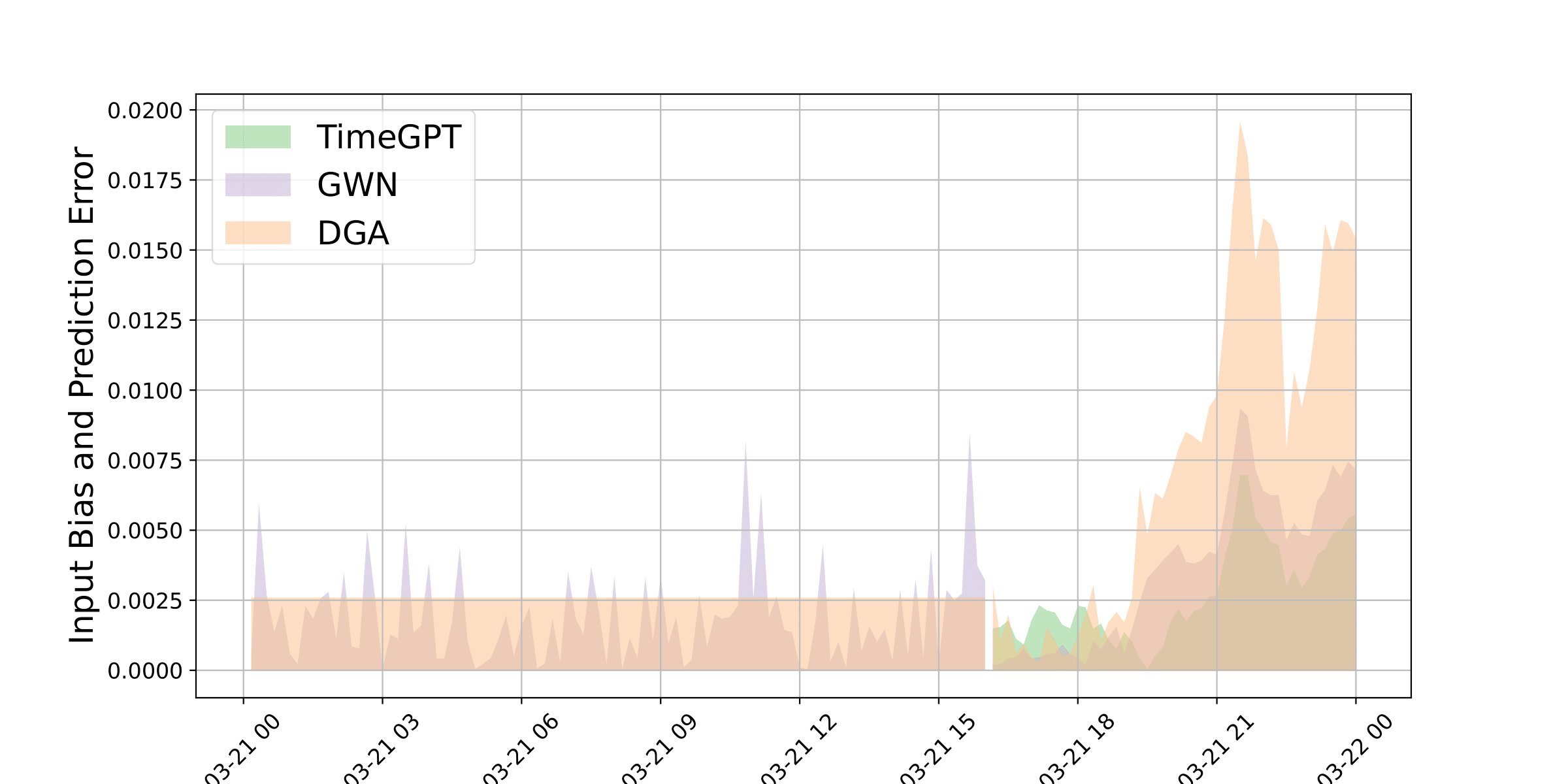}
}
\caption{(a) Inputs and predictions from LLMTime (using GPT-3.5) on the ETTh1 dataset; (b) Input bias and prediction errors corresponding to (a); (c) Inputs and predictions from LLMTime (using GPT-3.5) on the ETTh2 dataset; (d) Input bias and prediction errors corresponding to (c); (e) Inputs and predictions from TimeGPT on the weather dataset; (f) Input bias and prediction errors corresponding to (e). This figure highlights the greater disruption caused by DGA compared to GWN, showing significant deviations from the ground truth.}
\label{fig:lines}
\end{figure*}

\subsection{Overall Comparison}

\begin{figure*}[!h]
\centering
\subfigure[ETTh1, LLMTime w/GPT-3.5]{
    \centering
    \includegraphics[width = 0.31\textwidth]{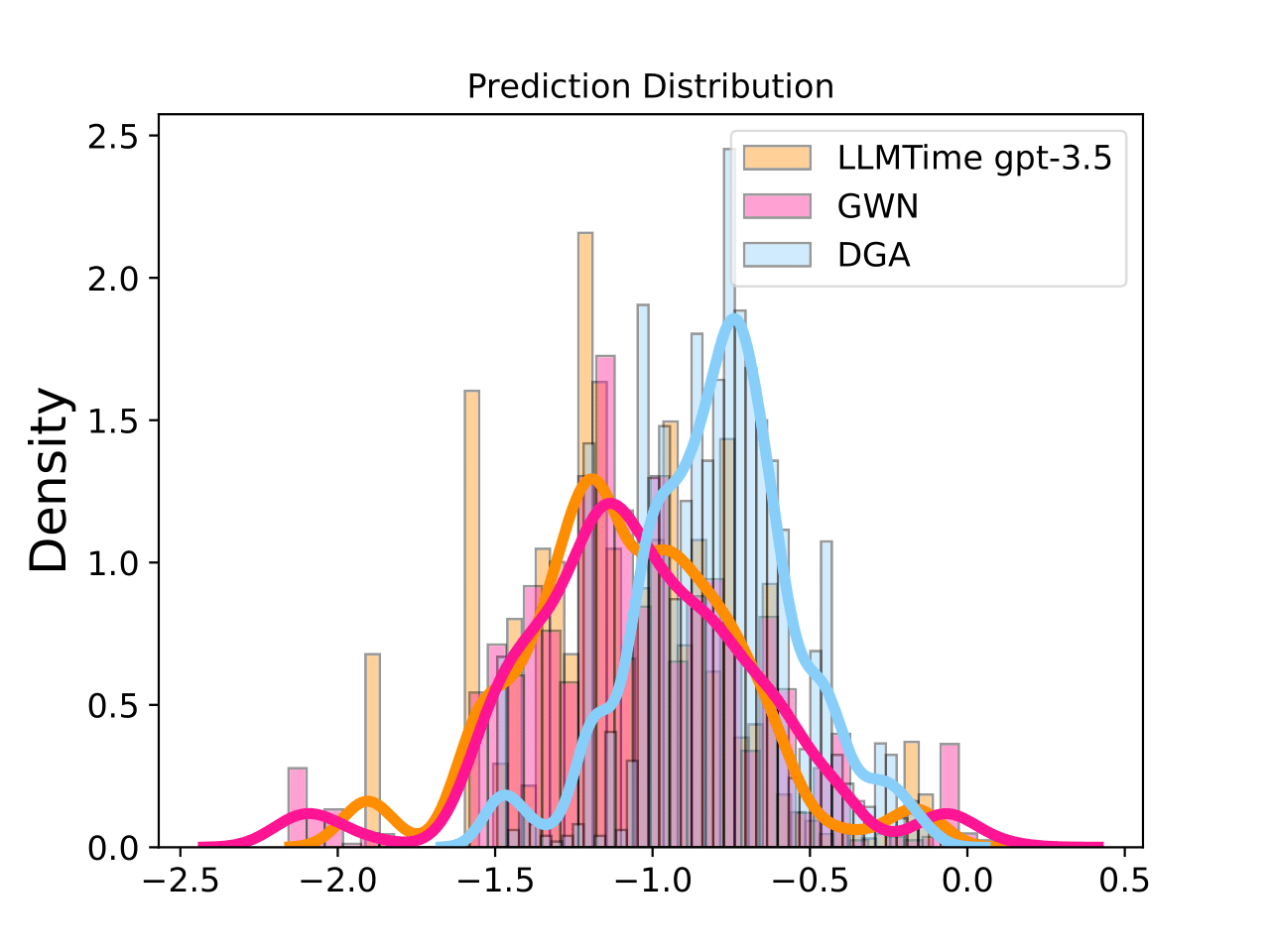}
}
\subfigure[ETTh2, LLMTime w/GPT-3.5]{
    \centering
    \includegraphics[width = 0.31\textwidth]{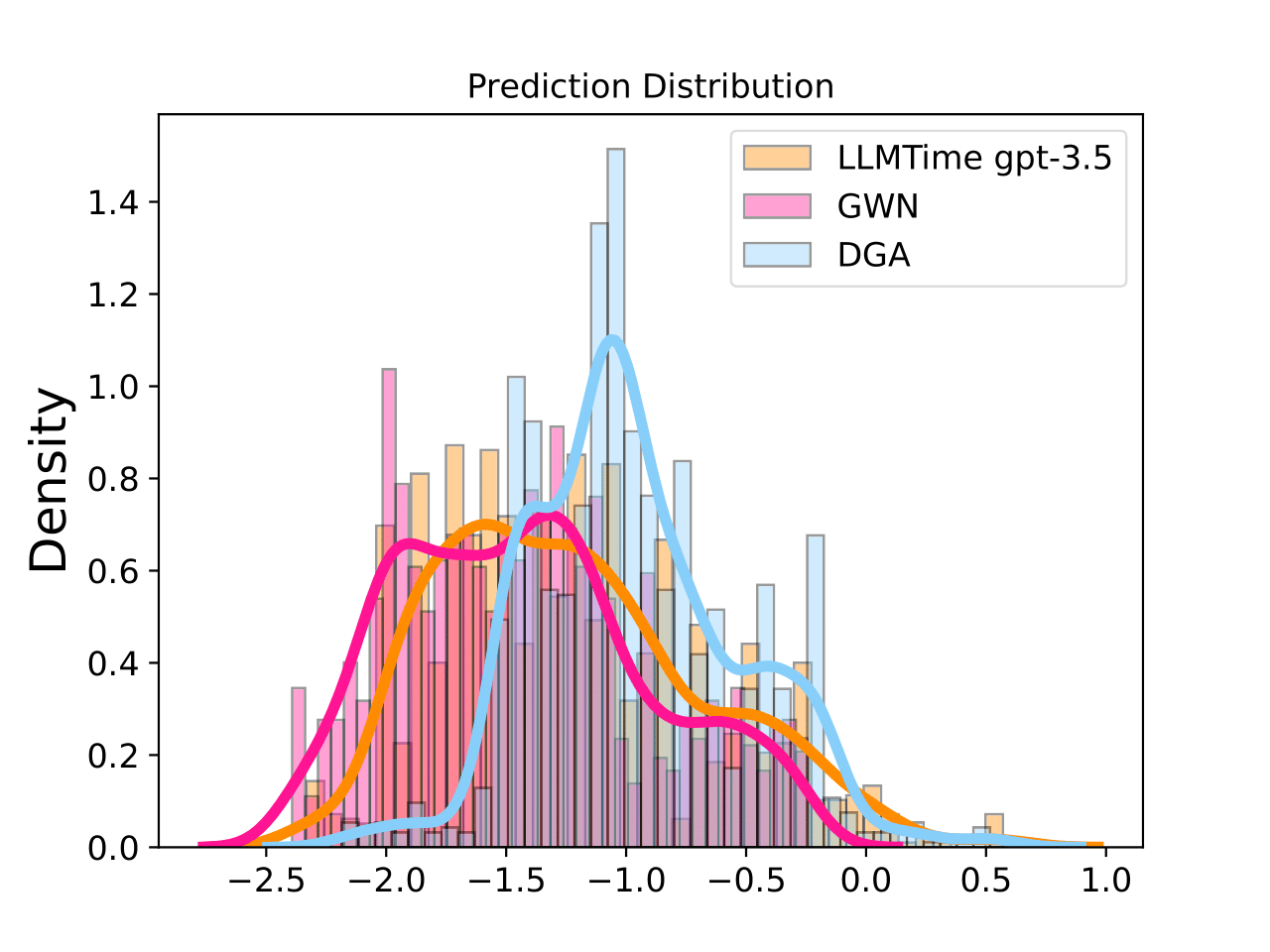}
}
\subfigure[Weather, LLMTime w/GPT-4]{
    \centering
    \includegraphics[width = 0.31\textwidth]{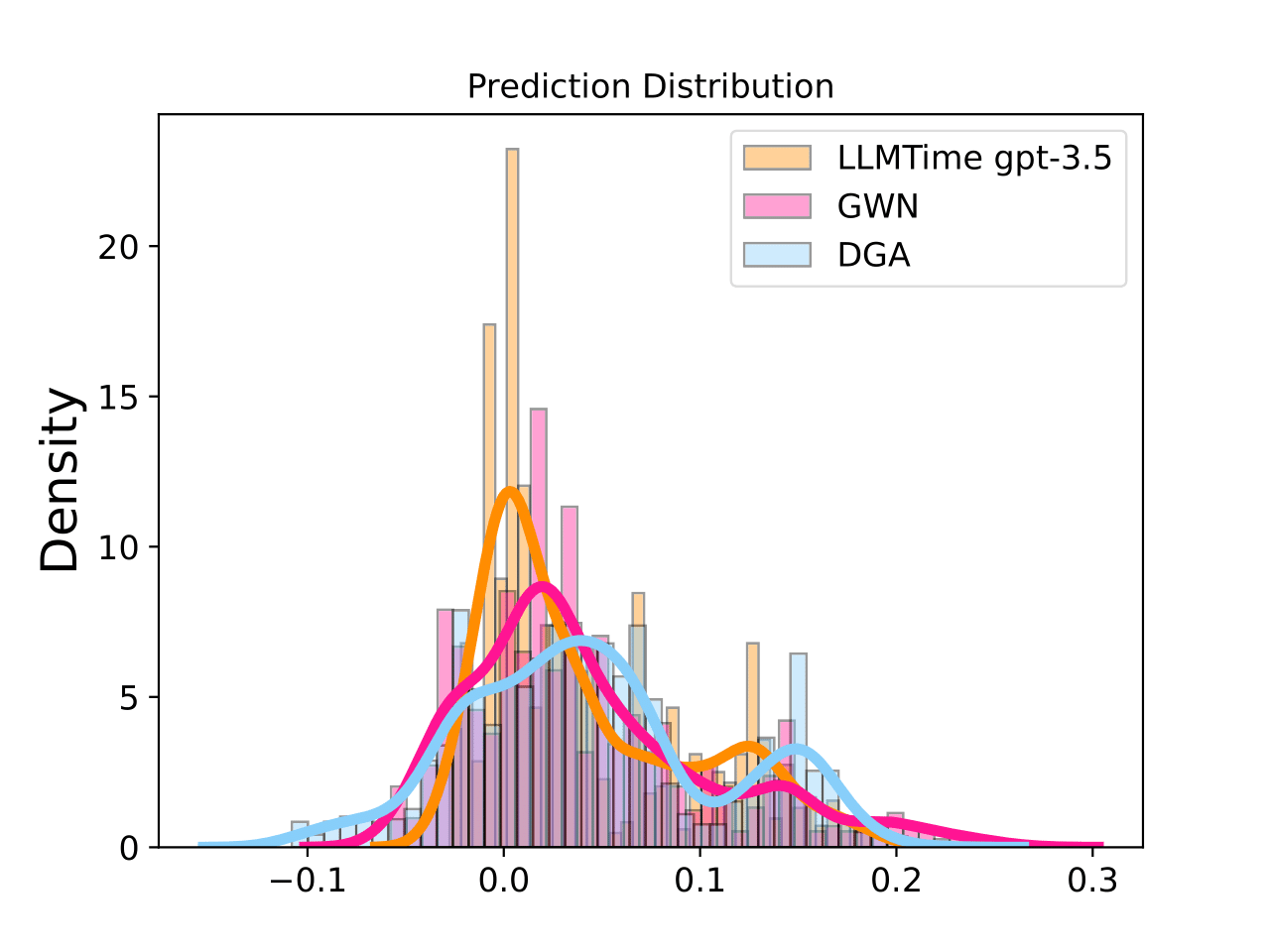}
}
\caption{Prediction distribution comparison for LLMTime (using GPT-3.5, GPT-4) across different datasets under clean input, GWN, and DGA.}
\label{fig:shift}
\end{figure*}
\begin{figure*}[!htbp] 
\centering
\subfigure[Clean Input]{
    \centering
    \includegraphics[width = 0.31\textwidth]{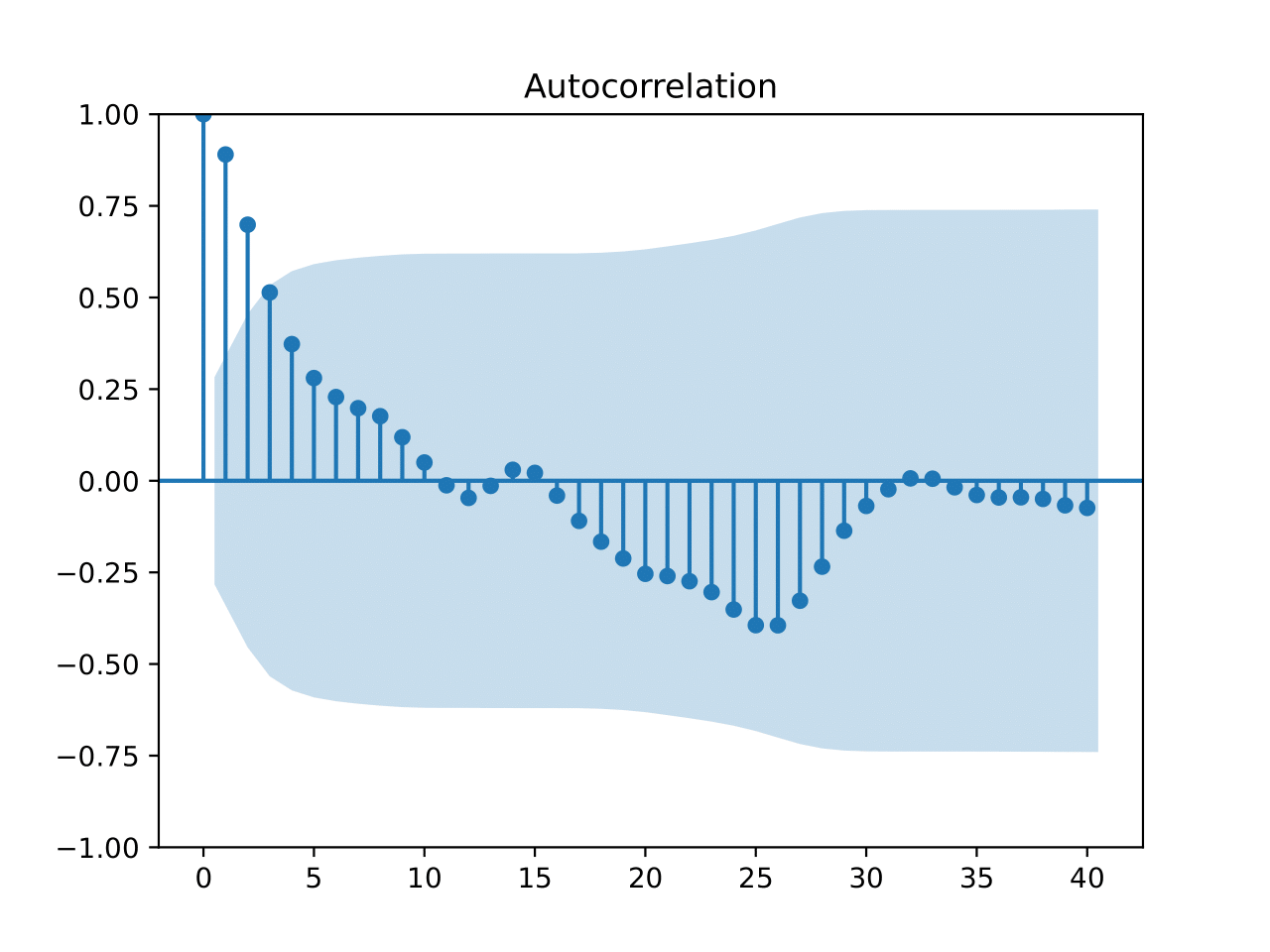}
}
\subfigure[Input with GWN]{
    \centering
    \includegraphics[width = 0.31\textwidth]{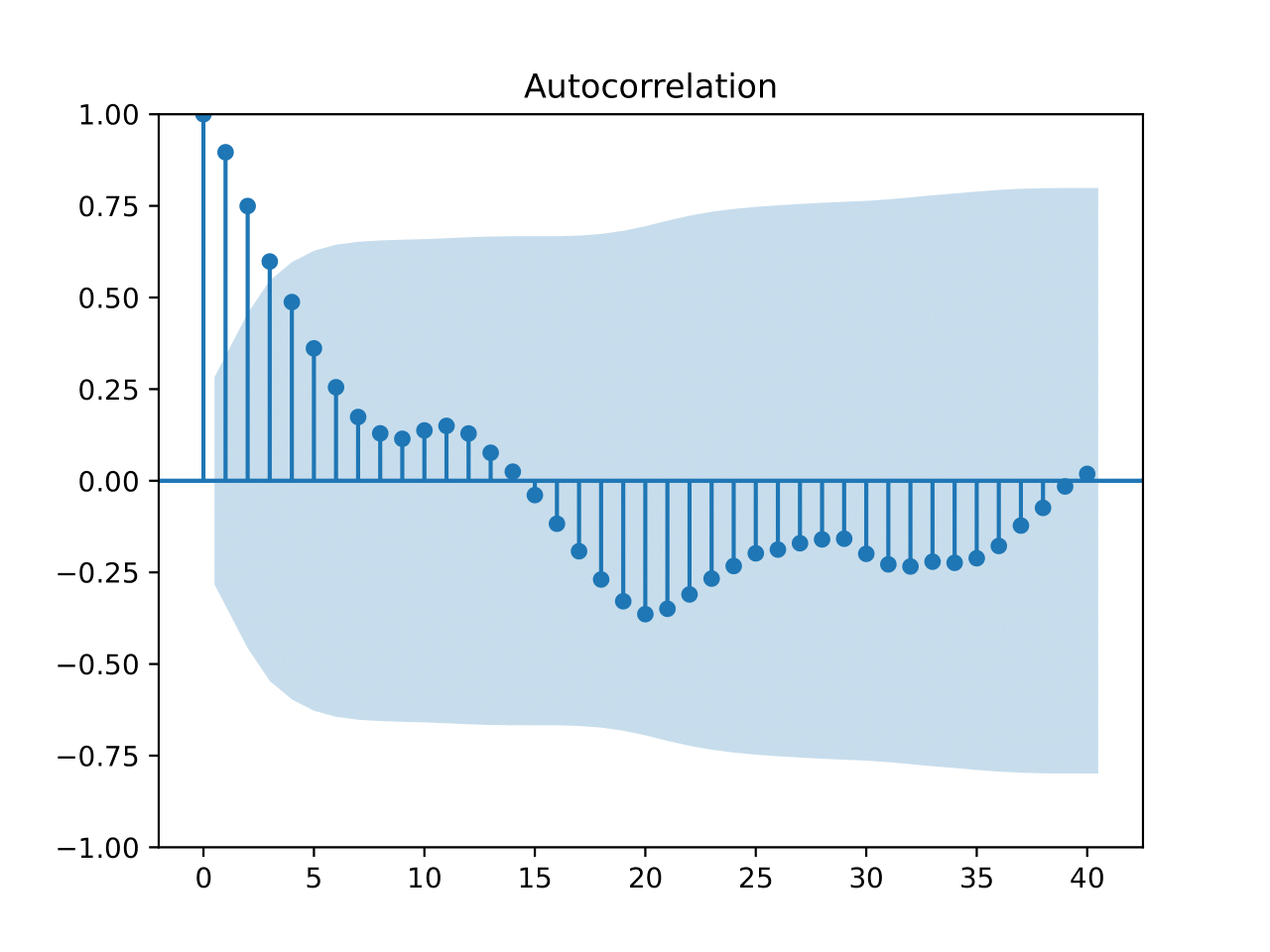}
}
\subfigure[Input with DGA]{
    \centering
    \includegraphics[width = 0.31\textwidth]{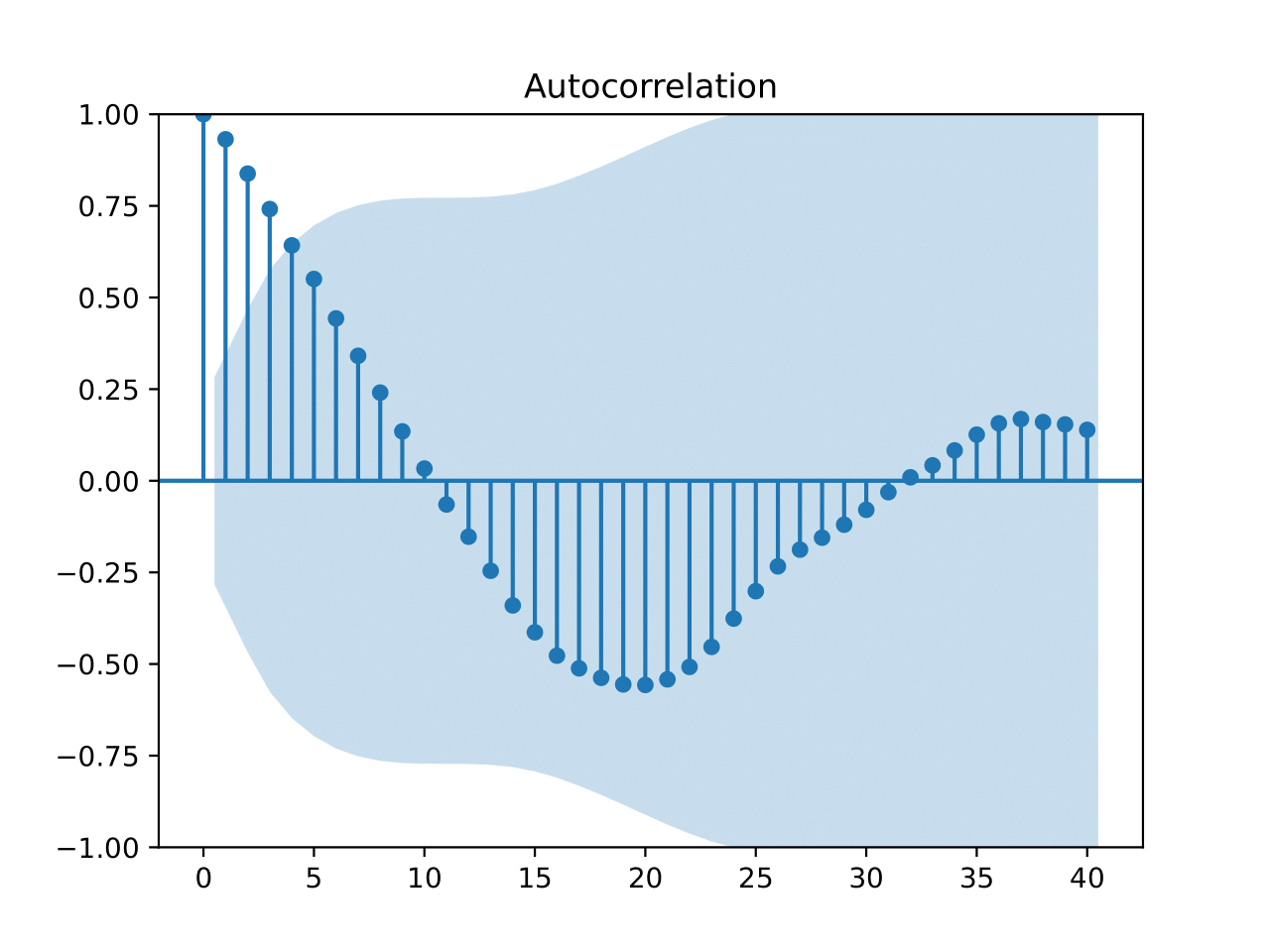}
}
\caption{Autocorrelation function curve comparison on ETTh2 by LLMTime using GPT-3.5 }
\label{fig:acf}
\end{figure*}

\begin{figure*}[!htbp] 
\centering
\subfigure[Weather, LLMTime w/GPT-4]{
    \centering
    \includegraphics[width = 0.31\textwidth]{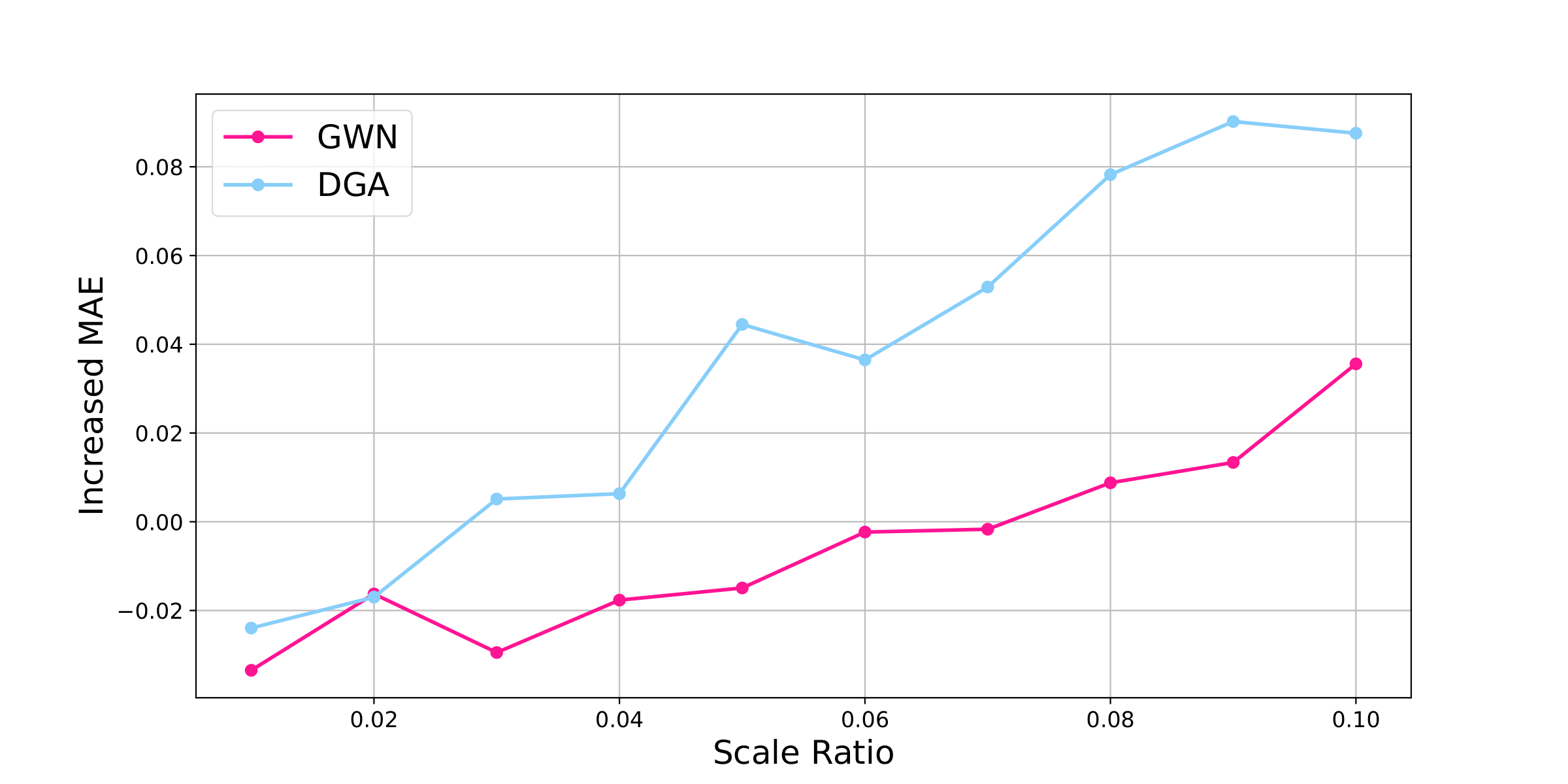}
}
\subfigure[Traffic, LLMTime w/Mistral]{
    \centering
    \includegraphics[width = 0.31\textwidth]{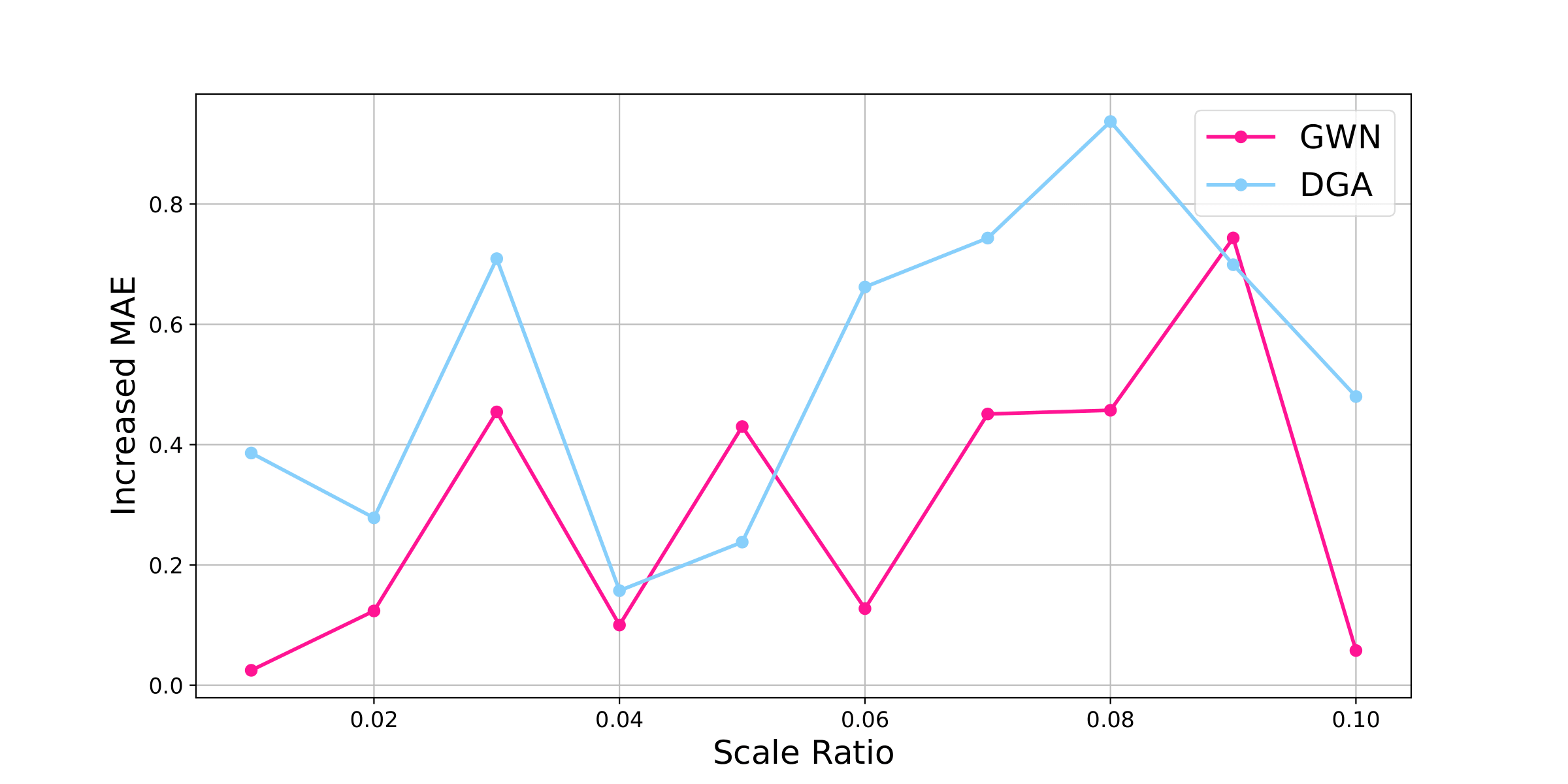}
}
\subfigure[Exchange, TimeGPT]{
    \centering
    \includegraphics[width = 0.31\textwidth]{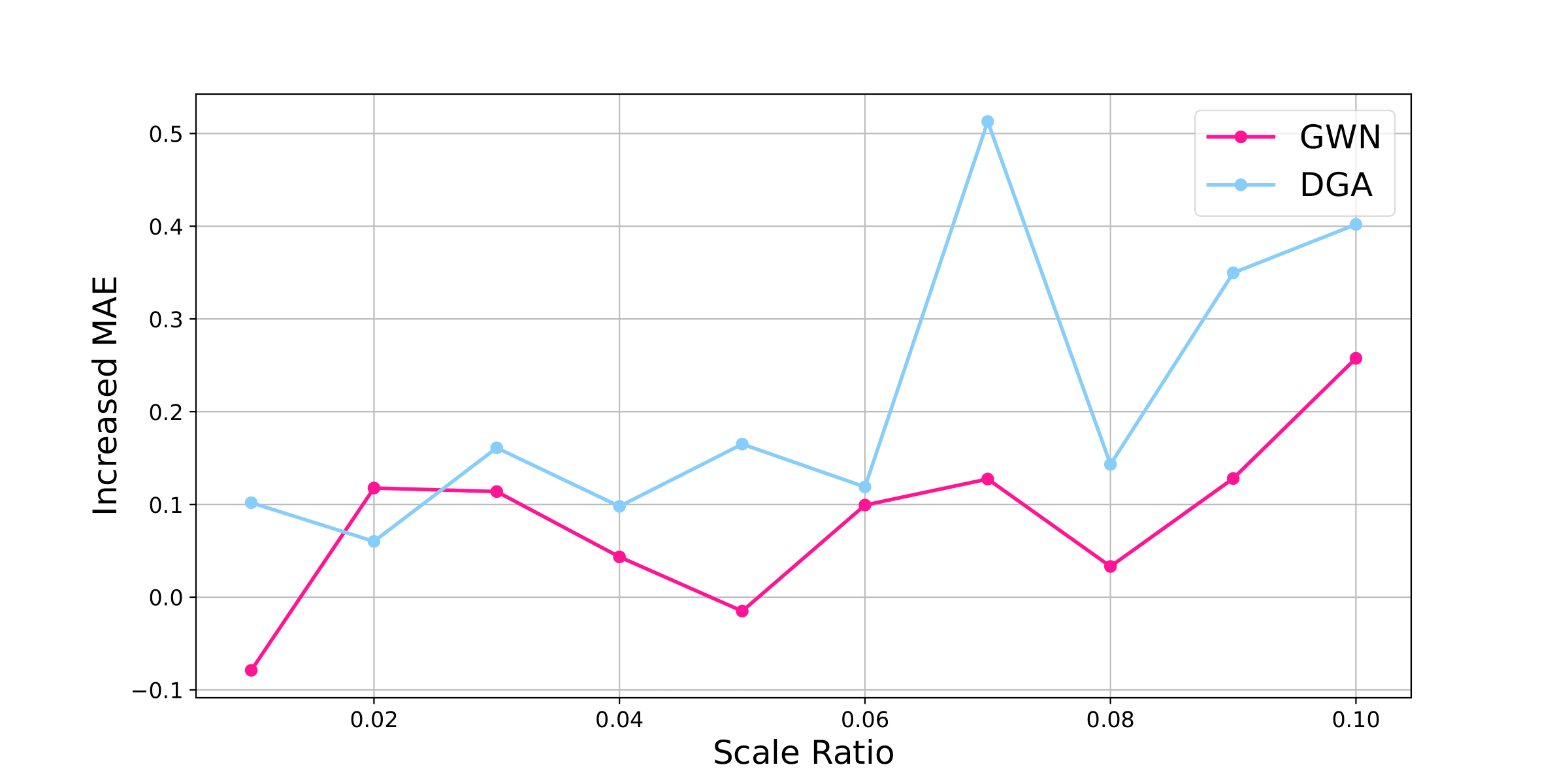}
}
\caption{Hyperparameter study on the effects of different scale ratios under GWN and DGA.}

\label{fig:hyper}
\end{figure*}

As shown in Table \ref{table:overall}, the experimental results demonstrate that the designed adversarial attacks significantly degraded forecasting performance across all datasets, as indicated by increased MSE and MAE values. Compared to GWN of the same perturbation intensity, our attacks had a much more detrimental effect on the models' predictions. For TimeGPT, which is pre-trained with large-scale time series data, the adversarial attack led to a sharp rise in forecasting errors, demonstrating that even models specifically built for time series forecasting are vulnerable. For LLMTime, which includes GPT-3.5, GPT-4, LLaMa, and Mistral as base models, the adversarial attack was even more pronounced. 

Figure~\ref{fig:robustness comparison} demonstrates the robustness comparison between LLM-based forecasting models (LLMTime with GPT-4 and TimeGPT) and non-LLM models (iTransformer and TimeNet) under the proposed black-box adversarial attack, DGA. The larger blue areas in the radar charts for the LLM-based models indicate that they experience significantly higher increases in errors, across all datasets (ETTh1, ETTh2, Exchange, Traffic, and Weather). In contrast, the non-LLM models, iTransformer and TimeNet, exhibit much smaller error increases, suggesting that they are more robust to adversarial attacks. This analysis highlights that LLM-based models are generally less resilient than non-LLM models, making them more vulnerable to adversarial manipulations in time series forecasting.

As illustrated in Figure \ref{fig:lines}, the attack caused a clear divergence between the forecasted values and the true time series, with all different variants of LLMTime exhibiting larger deviations compared to GWN. LLMTime with GPT-3.5 in particular, showed significant susceptibility, with their errors increasing substantially under adversarial conditions.


Across all models and datasets, the adversarial perturbations introduced significantly greater disruptions than GWN, clearly impacting the predictions and demonstrating the precision of the attack in destabilizing LLM-based forecasting. The magnitude of the degradation in predictive accuracy highlights the effectiveness of the proposed DGA. These findings emphasize the urgent need for robust defensive strategies to safeguard LLM-based forecasting models against adversarial threats. The current vulnerability of these models presents a significant challenge for real-world applications, particularly in high-stakes domains such as financial forecasting, energy demand prediction, and intelligent transportation systems. Without adequate defenses, adversarial attacks could lead to erroneous predictions, resulting in potential financial losses, inefficient resource allocations, or compromised safety in critical infrastructure.  

\subsection{Interpretation Study}
Figure \ref{fig:shift} illustrates the distribution shift in predictions caused by targeted perturbations on the LLM-based forecasting model. The proposed DGA method is designed to mislead the forecasting model, causing its predictions to resemble a random walk. As depicted in Figure \ref{fig:shift}, the "blue" shaded area, representing the perturbed prediction distribution, deviates significantly from the original "yellow" distribution and approaches a normal distribution. This shift underscores how subtle, well-crafted perturbations can manipulate the model into producing inaccurate forecasts. The effect of DGA-induced perturbations is pronounced when examining the prediction distributions, where errors are much more severe compared to the minor disruptions caused by GWN. These findings suggest that LLM-based forecasting models are highly susceptible to adversarial attacks that exploit the model's inherent vulnerabilities.

Additionally, the autocorrelation function (ACF) analysis provides further evidence of the detrimental impact of these adversarial attacks. Normally, LLMs demonstrate a strong ability to capture the temporal dependencies within time series data, maintaining coherent relationships between consecutive data points. However, as illustrated in Figure \ref{fig:acf}, when subjected to adversarial perturbations, these temporal dependencies break down, resulting in forecasts that no longer reflect the true underlying trends of the data. The disrupted autocorrelation patterns clearly illustrate the model's difficulty in preserving the natural flow of time series data under attack. In contrast, the addition of Gaussian noise, though introducing some fluctuations, does not cause the same level of disruption, maintaining a closer relationship to the clean data.

\subsection{Hyperparameter Study}

We systematically analyze the impact of varying scale ratios on model performance under both GWN and DGA adversarial perturbations. The vertical axis in Figure~\ref{fig:hyper} represents the increase in MAE, serving as a measure of the extent to which each attack degrades the precision of the forecast. This experiment is conducted across three distinct datasets using three LLM-based forecasting models.  

As illustrated in Figure~\ref{fig:hyper}, the DGA consistently induces a more pronounced increase in MAE compared to GWN as the scale ratio increases. This trend highlights the greater effectiveness of the DGA in destabilizing model predictions, as it is specifically designed to exploit model vulnerabilities rather than introduce random noise. To achieve a balance between imperceptibility and attack effectiveness, we determine that an optimal perturbation scale can be set at 2\% of the mean value of the given dataset. This choice ensures that the adversarial perturbation remains subtle enough to evade detection while still significantly impairing the forecasting model’s performance. 

\subsection{Discussion of Mitigation Methods}
\textbf{Challenges of Adversarial Training in LLM4TS}: Adversarial training effectively mitigates attacks but poses challenges for LLM-based time series forecasting due to the high computational costs of pretraining on large datasets. Its iterative nature further escalates costs, as it requires generating adversarial examples and optimizing against them during training. Retrofitting pre-trained models like TimeGPT with adversarial defenses is similarly impractical, often demanding months of computation and significant pipeline modifications.

\textbf{Alternative Mitigation Strategies for LLM4TS}: To overcome these challenges, preprocessing-based methods provide a practical alternative. Filter-based defenses reform time series data before forecasting, while machine learning-based anomaly detection identifies and filters adversarial inputs. These computationally efficient approaches are particularly effective against black-box adversarial attacks, offering a promising defense for LLM4TS models. Future work will focus on refining and evaluating these strategies.

\section{CONCLUSION}
In this study, we demonstrated the significant vulnerabilities of LLM-based models for time series forecasting to adversarial attacks. Through a comprehensive evaluation of TimeGPT and LLMTime (with GPT-3.5, GPT-4, LLaMa, and Mistral as base models), we found that targeted adversarial perturbations, generated using Directional Gradient Approximation (DGA), caused substantial increases in prediction errors. These attacks were far more damaging than Gaussian White Noise (GWN) of similar intensity, highlighting the precision and effectiveness of the adversarial strategy.

The experimental results revealed that both large, pre-trained models like TimeGPT and fine-tuned models such as LLMTime are highly susceptible to adversarial manipulation. The proposed attack can significantly degrade model performance across various datasets. This poses serious challenges for the deployment of LLMs in real-world time series applications, where reliability is critical.

Our findings emphasize the need for future research to focus on developing robust defense mechanisms to mitigate adversarial threats and enhance the resilience of LLM-based time series forecasting models. Without such protections, these models remain vulnerable to attacks that could undermine their practical utility in high-stakes environments. 

\newpage
\bibliographystyle{plainnat}
\bibliography{ref}

\end{document}